\documentclass[runningheads]{llncs}
\usepackage{graphicx}
\usepackage{comment}
\usepackage{amsmath,amssymb}
\usepackage{color}
\usepackage{microtype}
\usepackage{booktabs}
\usepackage{multirow}
\usepackage{hhline}
\usepackage{xspace}
\usepackage{epsfig}
\usepackage[dvipsnames]{xcolor}

\usepackage{graphicx}
\graphicspath{{./images/}}

\usepackage{overpic}
\usepackage{comment}

\usepackage{soul}
\usepackage{bm}

\usepackage{enumitem}

\usepackage[tableposition=top]{caption}
\usepackage[font=small,labelfont=bf, labelsep=period]{caption}

\usepackage[T1]{fontenc}
\usepackage[utf8]{inputenc}
\usepackage[font=small,labelfont=bf]{caption}
\newcommand{\field}[1]{\mathbb{#1}}

\newcommand{\R}{\field{R}}

\DeclareMathOperator*{\argmin}{arg\,min}

\newcommand{\bx}{\textbf{x}}

\newcommand{\bp}{\textbf{p}}

\newcommand{\by}{\textbf{y}}

\newcommand{\mS}{\mathcal{S}}
\newcommand{\mT}{\mathcal{T}}

\makeatletter
\DeclareRobustCommand\onedot{\futurelet\@let@token\@onedot}
\def\@onedot{\ifx\@let@token.\else.\null\fi\xspace}
 
\def\ie{\emph{i.e}\onedot}

\makeatother

\definecolor{darkgreen}{rgb}{0.0, 0.5, 0.0}

\begin{document}
\pagestyle{headings}
\mainmatter

\title{Joint Supervised and Self-Supervised Learning\\ for 3D Real-World Challenges}

\titlerunning{Joint Supervised and Self-Supervised Learning for 3D Real-World Challenges}

\author{Antonio Alliegro$^1$ \and
Davide Boscaini$^2$\and
Tatiana Tommasi$^1$}
\authorrunning{A. Alliegro et al.}

\institute{$^1$Politecnico di Torino, Italy \\
$^2$Fondazione Bruno Kessler, Trento, Italy \\
\email{\{antonio.alliegro,tatiana.tommasi\}@polito.it, d.boscaini@fbk.eu}}

\maketitle

\begin{abstract}
Point cloud processing and 3D shape understanding are very challenging tasks for which deep learning techniques have demonstrated great potentials.
Still further progresses are essential to allow artificial intelligent agents to interact with the real world, where the amount of annotated data may be limited and integrating new sources of knowledge becomes crucial to support autonomous learning.
Here we consider several possible scenarios involving synthetic and real-world point clouds where supervised learning fails due to data scarcity and large domain gaps.
We propose to enrich standard feature representations by leveraging self-supervision through a multi-task model that can solve a 3D puzzle while learning the main task of shape classification or part segmentation.
An extensive analysis investigating few-shot, transfer learning and cross-domain settings shows the effectiveness of our approach with state-of-the-art results for 3D shape classification and part segmentation.
\vspace{-3mm}
\end{abstract}

\section{Introduction}
\label{sec:intro}
Artificial intelligence made great strides in recent years with terrific results in the area of computer vision.
This naturally reflects also in our every-day life with apps able to collect images and automatically classify objects, detect faces, recognize places, and much more.
Still the task of fully understanding our real world remains far-fetched for many reasons, ranging from the inherent difficulty of dealing with a three-dimensional space, as well as time and domain variations which makes it difficult to learn robust models able to generalize to any new scenario.
Currently, many of those issues are the same that maintain a large gap between having a smartphone in our hands and a robot at home.
Embodied intelligent systems need more than 2D visual perception: 3D shapes have to be reliably recognized regardless of the plethora of environmental constraints, and possibly segmented in their functional parts to allow a robot completing a simple tasks as can be opening a honey jar.

Thanks to the rise of powerful computational resources, 3D research is also progressively flourishing together with new ways to collect and describe 3D data.
LiDAR scanners and stereo cameras gave rise to massive point cloud datasets possibly spanning even large entities such as an entire city.
However, they come with three main drawbacks: point clouds are un-structured, un-ordered and eager for precise manual annotation due to many possible sources of noise.
The first two properties make typical convolutional neural networks (CNN) unsuitable for point clouds.
\begin{figure*}[t!]
    \centering
    \includegraphics[width=\linewidth]{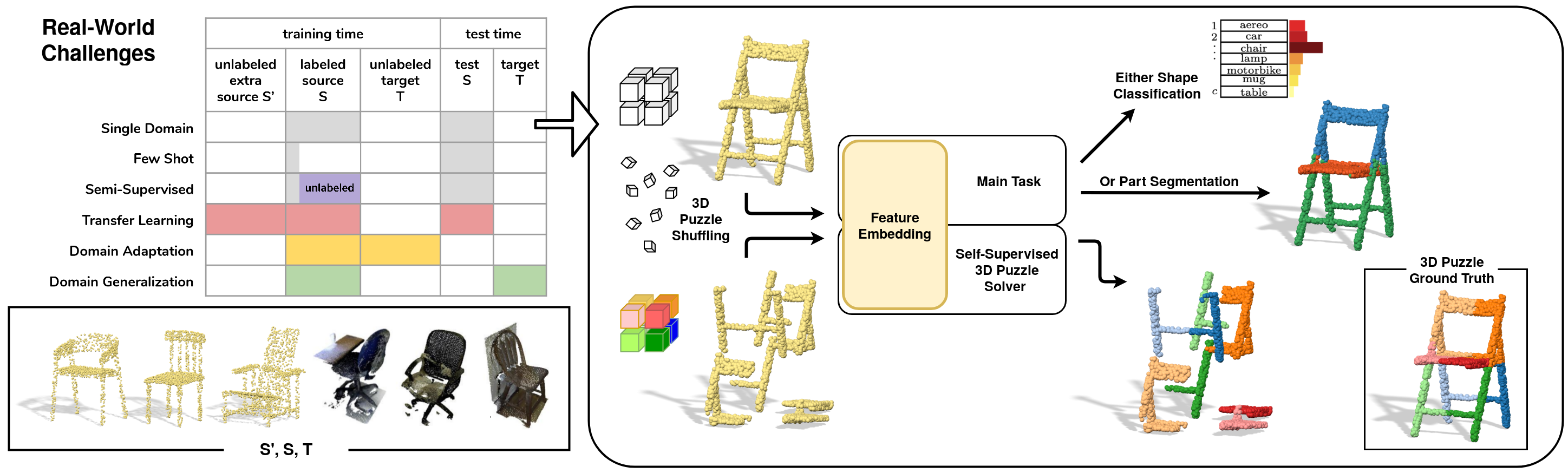}
    \caption{Overview of the considered real world challenges and of the proposed multi-task approach. We deal with different domains covering real and synthetic 3D point clouds as well as several learning settings across domains and scarce annotations}
    \label{fig:teaser}
    \vspace{-0.6cm}
\end{figure*}
Possible solutions consists in rendering point clouds to multiple 2D views or pre-processing them with a voxelization procedure to make data suitable for 3D CNN, but these techniques are either computationally expensive or come with inevitable loss of information and with negative effects on the overall recognition or segmentation performance.
The third property has initially guided research towards very well lab-controlled and synthetic CAD object datasets where labeling is simpler.
However, the most recent results on those kind of testbed are witnessing a trend of performance saturation raising the question of how to move forward.
All these challenges describe a research area in need of new 
deep learning models able to deal with large amount of unsupervised real-world point cloud data.\\ 

\noindent We can summarize the contributions of our work as following:\\
    \indent 1. we design a new research landscape by 
    \emph{tackling at once several aspects of cross-domain adaptive learning for 3D vision in real world scenarios.} 
    Recent 2D image analysis literature has shown how the lack of data annotation and the possible domain shift issues can be alleviated by integrating different source of knowledge in the learning process through Transfer Learning (TL), Domain Adaptation (DA) and Generalization (DG).
    Self-supervised learning has also shown to be a helpful support (see Section~\ref{sec:related} for an overview). We investigate how to follow this trend for 3D tasks. \\
    \indent 2. we propose a \emph{new multi-task end-to-end deep learning model for point clouds that combines supervised and self-supervised learning} 
    (see Fig.~\ref{fig:teaser}). Specifically we define a deep architecture that solves 3D puzzles while jointly training a main recognition task. We show how these two tasks complement each other making the obtained model (a) more precise in case of large amount of labeled data, (b) more robust in case of scarce labeled data, (c) easier to transfer for adaptation and (d) more reliable for out of domain generalization.\\
    \indent 3. we present extensive experiments across three different point clouds datasets: our multi-task method outperforms the standard supervised learning baseline and defines the new state-of-the-art for both shape classification and part segmentation in the most challenging real world settings.

\section{Related Work} \label{sec:related}
How to use powerful deep learning methods for \emph{supervised learning} of classification and segmentation models on point clouds is an extremely active area of research.
Early approaches roots back to PointNet \cite{pointnet_CVPR17}, a MLP architecture combining symmetry and spatial transform functions to learn point features which are then aggregated into global shape representations.
PointNet++ \cite{pointnet++_NIPS2017} extended the previous model by hierarchically combining multiple PointNet modules.
PointCNN \cite{NIPS2018_7362} maps shape vertices to a canonical space where their order is preserved and therefore allows the application of traditional convolutional operators on them.
Many other solutions have also been proposed with the aim of extending convolutional filters on point clouds either in the spatial \cite{monet_2017,xu2018spidercnn,ben20183dmfv,dgcnn} or in the spectral domain \cite{lscnn_2015,localspectral,Yi_2017_CVPR}.

\emph{Self-supervised learning} is a framework recently achieving large attention in the 2D computer vision community.
It deals with originally unlabeled data for which a supervised signal is obtained by first hiding part of the available information and then trying to recover it.
This procedure is generally indicated as \emph{pretext} task and possible examples are image completion \cite{pathakCVPR16context}, colorization \cite{zhang2016colorful,Larsson_2017_CVPR}, patch reordering \cite{doersch2015unsupervised,noroozi2016} and rotation recognition \cite{gidaris2018unsupervised}.
The solution of the pretext task captures high-level semantic knowledge from the data so that the learned representation can be transferred to other \emph{downstream} tasks as a powerful warm-up initialization.
Self-supervision has shown to be relevant also to describe 3D structures. Recent works proposed autoencoder-based approaches to reconstruct 3D point clouds \cite{sonet_2018_CVPR,Zhao_2019_CVPR} and methods to deform a 2D grid onto the underlying 3D object surface \cite{Yang_2018_CVPR}.
In \cite{Han_2019_ICCV} point clouds are split into a front and a back half from several angles and a model is trained to predict one from the other.
A network to verify whether two randomly sampled parts from the dataset belong or not to the same object is presented in \cite{3dworkshopcvpr}, while \cite{saudersievers} proposes to reconstruct point clouds whose parts have been randomly rearranged.
Finally, reconstruction, clustering, and self-supervised classification are combined together in \cite{Hassani_2019_ICCV}, defining a fully unsupervised multi-task approach for feature learning.

The pretext and downstream stages define a particular case of \emph{Transfer Learning} where unsupervised data is exploited to support supervised learning on a task of interest.
In many real world applications, despite unlabeled data may be freely available, their distribution can significantly differ from that of the supervised data at hand, rising the extra issue of how to deal with a \emph{domain shift} for which the transfer procedure may backfire.
A similar problem holds also when the unsupervised collection is not an extra source of knowledge, but corresponds instead to the test on which we need to evaluate a supervised model.
For instance, when train and test data are respectively drawn from photos and painting or from virtual reality simulators and real world pictures.
\emph{Domain Adaptation} literature focuses on this scenario supposing that the unsupervised test data is transductively available at training time.
Many adaptive solutions have been proposed in the last years for 2D vision problems involving either feature alignment strategies with dedicated losses \cite{Long:2015,LongZ0J17,dcoral,hdivergence}, ad-hoc network layers \cite{mancini2018boosting,carlucci2017auto} or adversarial learning \cite{Ganin:DANN:JMLR16,Hoffman:Adda:CVPR17}.
Several recent works also involve generative style transfer methods \cite{russo17sbadagan,cycada}, reconstruction penalties \cite{DGautoencoders,Li_2018_CVPR,Bousmalis:DSN:NIPS16} or feature norms constraints \cite{Xu_2019_ICCV}.
An even wider and more challenging problem is the one tackled by \emph{Domain Generalization}.
In this case, the specific target data are not provided at training time and the goal is that of learning a model robust to any kind of new domain shift that can appear during deployment.
Only few works have shown good results in this setting, mainly considering feature alignment among multiple data sources when available \cite{DGautoencoders,Li_2018_CVPR,Li_2018_ECCV,ADAGE19}, data augmentation \cite{DG_ICLR18,Volpi_2018_NIPS}, and meta-learning \cite{MLDG_AAA18,Li_2019_ICCV}.
Most recently, self-supervised learning has also shown promising results in the DA and DG scenarios \cite{Carlucci_2019_CVPR,lopez,Zhai_2019_ICCV}.

\emph{Our approach connects all the described frameworks in a novel way}. Instead of using one \cite{Han_2019_ICCV,saudersievers} or multiple \cite{3dworkshopcvpr,Hassani_2019_ICCV} self-supervised tasks as pretext, we consider self-supervision as an auxiliary objective to be optimized jointly with the supervised one in a multi-task model.
Specifically, we choose to define and solve 3D puzzles while learning to classify or segment 3D shapes. We focus on real word point clouds which may be severely cluttered with noise and background points \cite{scanobjectnn_ICCV19}. Our analysis largely extends that recently presented in \cite{pointdan10} which has just started to investigate DA for 3D point clouds and does not tackle DG nor TL and other challenging settings like semi-supervised learning.

\section{Method} \label{sec:method}
\paragraph{The Intuition.} 
At the basis of our work there is the intuition that 3D point cloud understanding can still be extremely challenging even when supervised knowledge is provided. This becomes particularly true when moving from synthetic to real-world data and tasks.
Due to their un-ordered nature, how to properly exploit local neighbours and at the same time taking into account the global structure of the 3D shape is quite challenging.
Self-supervised learning is helpful in this respect: a simple task like solving a 3D puzzle leverages on the spatial co-location of shape parts and exploits reliable knowledge on relative point positions both at global and local level.
Thus, while learning to solve a 3D puzzle we gain useful knowledge that can support recognition at different scales (whole object and parts).
We design our learning model as a multi-task deep network where a main recognition task and the self-supervised puzzle task jointly learn a shared data representation.

\begin{figure}[ht!]
    \centering
    \includegraphics[width=\linewidth]{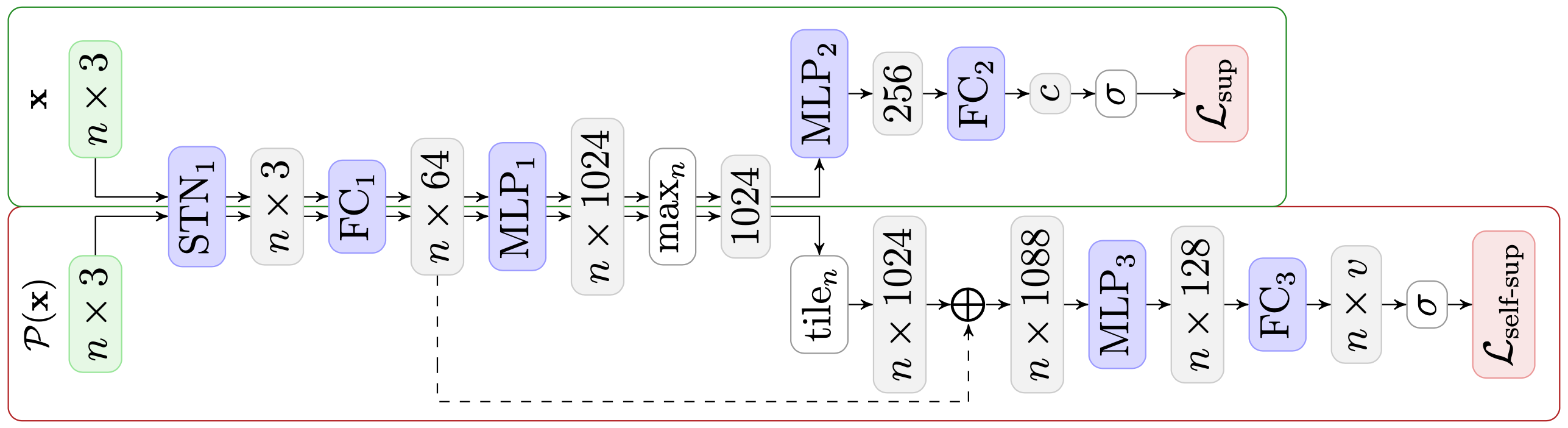}
    \includegraphics[width=\linewidth]{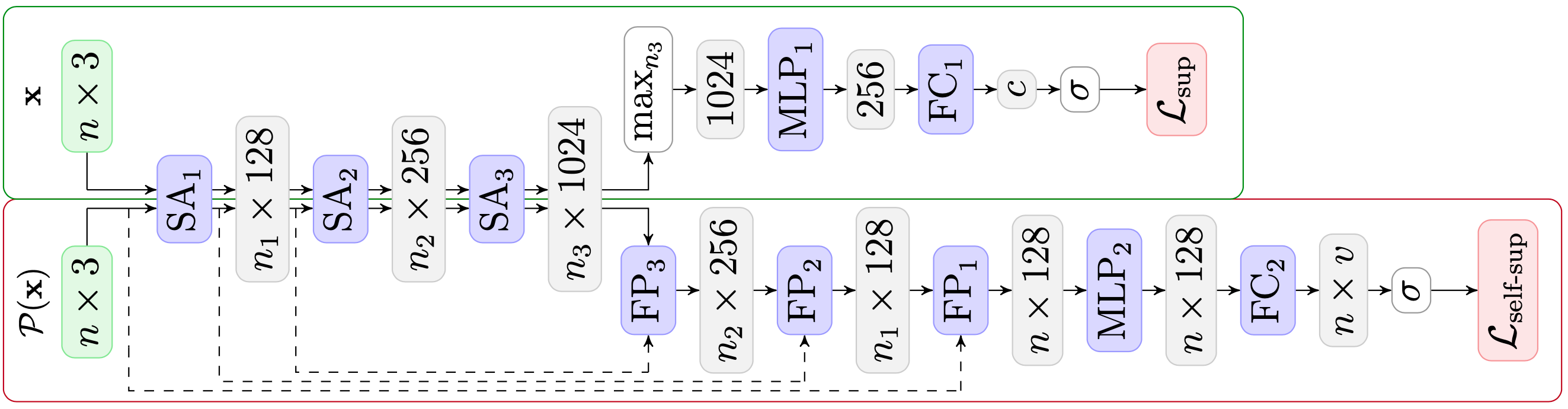}
\caption{Our multi-task architecture with PN (top) and PN++ (bottom) backbone used for shape classification. We refer the interested reader respectively to \cite{pointnet_CVPR17} and \cite{pointnet++_NIPS2017} for the details of each component. Color scheme: green = input data, blue = parametric layer, white = non-parametric layer, grey = output features, red = loss function}
\label{fig:arch_pn}
\vspace{-6mm}
\end{figure}

\vspace{-2mm}\paragraph{More Formally.} 
Let us assume to observe data $S=\{(\bx_i,\by_i)\}_{i=1}^N$ where each sample $\bx_i=\{\bp_1^i,\bp_2^i,\ldots,\bp_K^i\}$ is a un-ordered set of $K$ 3D points $\bp_k^i\in \R^3$ described by their Euclidean coordinates. The corresponding label $\by_i$ depends on the specific task at hand.
In case of \emph{shape classification}, $\by_i=y_i\in\{1,\ldots,C\}$ is a scalar denoting one out of $C$ object categories.
For \emph{part segmentation}, instead, $\by_i$ is a $K$-dimensional vector with each component $y_{ik}\in\{1,\ldots,Q\}$ where $Q$ is the number of object parts.
In the following we will refer either to classification or part segmentation as our \emph{main task} and we will describe how each of them can be jointly learned with the auxiliary self-supervised task of 3D puzzle solving.
Our multi-task model can be described as the combination of two parametric non-linear functions:
$\mathbf{\Phi}_{\theta_f,\theta_m}$ and $\mathbf{\Psi}_{\theta_f,\theta_p}$, where the subscripts of the parameters $\theta$ refer respectively to the feature extraction ($f$), main task ($m$), and puzzle solution ($p$) modules of our deep network.
The feature encoder is shared between the two functions and is in charge of summarizing the local and global geometric information from the input point cloud to a richer latent space.
For each sample $\bx$ that enters the network,  $\mathbf{\Phi}_{\theta_f, \theta_m}(\bx)$ is the output of the feature extractor and final fully connected part of the network. It is finally compared to its corresponding ground-truth label through the loss function $\mathcal{L}_m(\mathbf{\Phi}_{\theta_f, \theta_m}(\bx), \by)$ which measures the prediction error on the main task.

The auxiliary function $\mathbf{\Psi}$ deals with a \emph{puzzled} variant $\tilde{\bx} = \mathcal{P}(\bx)$ of the original input point cloud.
To get it, we start from $\bx$, scale it to unit cube and split each axis into $l$ equal lengths intervals forming $l^3$ voxels which are labeled according to their original position. Each vertex contained inside a voxel inherits its label.
Finally, all the voxels are randomly swapped, producing a new shuffled point cloud.
We indicate with $\tilde{S}=\{(\tilde{\bx}_i,\tilde{\by}_i)\}_{i=1}^N$ the obtained puzzled samples where the voxel position label for each point is $y_{ik} \in \{1,\ldots,l^3\}$.
Once these new displaced data are encoded in the feature latent space, a second network head focuses on solving the 3D puzzle problem by minimizing the auxiliary loss that measures the reordering error $\mathcal{L}_p(\mathbf{\Psi}_{\theta_f, \theta_p}(\tilde{\bx}),\tilde{\by})$ in terms of difference between the assigned voxel label and the correct one per point.
Overall we train the network to obtain the optimal model through
\begin{equation}
    \argmin_{\theta_f, \theta_m, \theta_p} \sum_{i=1}^N \mathcal{L}_m(\mathbf{\Phi}_{\theta_f, \theta_m}(\bx_i),\by_i) + \alpha\, \mathcal{L}_p(\mathbf{\Psi}_{\theta_f, \theta_p}(\tilde{\bx}_i), \tilde{\by}_i), 
\end{equation}
where both $\mathcal{L}_m$ and $\mathcal{L}_p$ are cross-entropy losses. Note that, while the first loss deals only with original samples, the second involve both original and puzzled samples, given the random nature of the voxel shuffling procedure.

\vspace{-2mm}\paragraph{Hyper-parameters and Implementation Choices.}
The described learning problem has one main hyper-parameters $\alpha$, which weights the self-supervised loss and balances its importance with respect to the main supervised task.
Since we exploit self-supervision as an auxiliary objective we reasonably assign less importance to it with respect to the main task and set $\alpha=0.6$ for all our analysis.
A further parameter of the problem is the axis quantization step used to define the puzzle parts: we set $l=3$. An ablation analysis on both $\alpha$ and $l$ is provided in sec. \ref{subsection:ablation}.
To realize our model we built over two well known and reliable architectures: PointNet (PN) and PointNet++ (PN++) by extending their structure with the inclusion of a new ending branch dedicated to 3D puzzle resolution.
Figure \ref{fig:arch_pn} illustrates the corresponding architectures for our multi-task approach.
By investigating both of these backbones we can highlight the effect of the context information learned by the puzzle to different ways of dealing with point clouds.
Indeed the first architecture basically learn on each point independently and only accumulates the final features, while the second follows a multi-scale strategy with a heuristic point grouping at separate layers.

\section{Experiments}
\label{sec:exper}

\subsection{Settings}
We consider several experimental settings involving a source dataset $\mS$ divided into two disjoint parts $\mS_{\text{train}}$, $\mS_{\text{test}}$, a possible extra set of unlabeled data from a different source domain $\mS'$ and an unlabeled target domain $\mT$, different from both $\mS$ and $\mS'$. We list them below:

\vspace{-3mm}\paragraph{Single Domain} (SD): the whole set of annotated samples from $\mathcal{S}_{\text{train}}$ are available for supervised learning.
We test on the portion $\mathcal{S}_{\text{test}}$ of the same original dataset.

\vspace{-3mm}\paragraph{Few-Shot} (FS): it considers the case of limited training samples. We reduce the cardinality of $\mS_{\text{train}}$ at different percentage scales and we evaluate on $\mS_{\text{test}}$.

\vspace{-3mm}\paragraph{Semi-Supervised} (SS): this setting is analogous to the previous one but the percentage of samples which is not included in $\mS_{\text{train}}$ can still be used as unlabeled data during training.

\vspace{-3mm}\paragraph{Transfer Learning} (TL): besides the annotated data from $\mS$, a further set of unlabeled samples from a different domain $\mS'$ is available at training time. In this case, when running the multi-task approach, we feed the self-supervised task with the extra unlabeled samples while the supervised data is only used for the main task. The final evaluation is performed on $\mS_{\text{test}}$.

\vspace{-3mm}\paragraph{Domain Generalization} (DG): this setting is analogous to SD but differs in the evaluation phase, where the performance is computed on the target collection $\mT$ (belonging to a different domain).

\vspace{-3mm}\paragraph{Domain Adaptation} (DA): both the supervised data $\mS$ and the unsupervised target data $\mT$ are available at training time and enter the self-supervised part of our multi-task method. Instead the main task is learned only on the supervised $\mS$.
As standard practice, the final model is evaluated on $\mT$ data.

\subsection{Datasets}
\paragraph{Synthetic data from ModelNet:}
ModelNet40~\cite{mn40_dataset} contains 12311 3D CAD models from 40 man-made object categories. We use the official dataset split, consisting of 9843 train and 2468 test shapes. By following \cite{pointnet_CVPR17}, from each CAD model we extract a point cloud by uniformly sampling 2048 vertices from the faces of the synthetic mesh. Each point cloud is then centered in the origin and scaled to fit in the unit sphere.

\vspace{-2mm}\paragraph{Synthetic data from ShapeNet:}
ShapeNet is one of the largest repositories of annotated 3D models. We use two variants of it depending on the annotations required by the main task we aim to solve.
ShapeNetCore~\cite{shapenet_dataset} contains 51300 clean 3D CAD models from 55 different classes, each annotated with the object category.
ShapeNetPart~\cite{shapenetpart_dataset} contains 16881 3D shapes from 16 different categories. We use the official dataset split containing 12137 train, 1870 validation, and 2874 test shapes. Each shape is annotated with 2 to 6 parts, for a total of 50 distinct parts among all categories. To reduce the high variability of vertex density across different categories, 2048 vertices are randomly sampled from each shape. Also in this case each point cloud is normalized to fit in the unit sphere.

\vspace{-2mm}\paragraph{Real-world data from ScanObjectNN:}
ScanObjectNN~\cite{scanobjectnn_ICCV19} is a recent dataset containing 2902 3D scans of real-world objects from 15 categories (mostly furniture) originated from ScanNet~\cite{dai2017scannet}. Real-world 3D scans are much more challenging than CAD models due to the presence of acquisition artifacts such as vertex noise, non-uniform vertex density, missing parts, and occlusions. Moreover, real-world data contains background vertices, which are absent in the synthetic models of ModelNet and ShapeNet.
Several variants of the ScanObjectNN dataset are provided. The vanilla version \texttt{OBJ\_ONLY} is the closest to synthetic datasets since it contains only foreground vertices. \texttt{OBJ\_BG} contains the same shapes but with the addition of background vertices. Finally, there are the most challenging cases where 50\% Translation, Rotation around the gravity axis and Scaling along each axis are applied to 3D scans: \texttt{PB\_T50\_RS\_BG} and \texttt{PB\_T50\_RS}, respectively with and without background.
Interestingly, 11 among the 15 categories of ScanObjectNN overlap with the those in ModelNet40, making it suitable for investigating the domain shift between the two datasets in terms of Domain Generalization (DG) and Domain Adaptation (DA) experiments.


\begin{table*}[t!]
    \centering
    \caption{Shape classification accuracy (\%) of our multi-task approach with respect to the main classification baseline $(\alpha=0)$ implemented on two different backbones (PN \cite{pointnet_CVPR17}, PN++ \cite{pointnet++_NIPS2017}). In the Transfer Learning (TL) setting, extra unsupervised data sources are integrated in the learning process as input to the self-supervised task (ShapeNet for ModelNet40 and ModelNet40 for ScanObjectNN). Suffix \texttt{BG} indicates that the point clouds contains background vertices}
    \resizebox{\textwidth}{!}{
        \begin{tabular}{l@{~~}|@{~~}l@{~~}|@{~~}c@{~~}|@{~~}c@{~~}c@{~~}c@{~~}c}
            \hline
            \multirow{2}{*}{Backbone} &
            \multirow{2}{*}{Method} &
            \multirow{2}{*}{ModelNet40} &
            \multicolumn{4}{c}{ScanObjectNN} \\
            & & & \texttt{OBJ\_ONLY} & \texttt{OBJ\_BG} & \texttt{PB\_T50\_RS} & \texttt{PB\_T50\_RS\_BG} \\
            \hline
            \multirow{3}{*}{PN } & Baseline & 88.65 & 75.22 & 70.22 & 71.37 & 62.56 \\
                & Our SD & 89.71 & 75.04 & \textbf{71.26} & 73.39 & 65.20 \\
                & Our TL & \textbf{90.72} & \textbf{77.45} & \textbf{71.26} & \textbf{73.49} & \textbf{65.61} \\
            \hline
            \multirow{3}{*}{PN++ } & Baseline & 91.93 & 84.17 & 83.99 & 78.66 & 77.90 \\
                & Our SD & \textbf{92.10} & \textbf{85.89} & 83.13 & 79.22 & 78.00 \\
                & Our TL & 91.58 & 84.68 & \textbf{84.33} & \textbf{80.46} & \textbf{79.08} \\
            \hline
        \end{tabular}
    }
    \label{tab:shape_class}
    \vspace{-5mm}
\end{table*}

\subsection{Classification Results}\label{subsection:classification}
We dedicate the first part of our experimental analysis to the main task of shape classification. The goal is to predict the object category of the observed point cloud from a set of $C$ classes. We evaluate the performance in terms of overall accuracy. All reported results are averaged over three runs.

\vspace{-3mm}\paragraph{Training details.} Throughout all the classification experiments we used the following parameters.
When using the PN 
backbone, we considered a batch size of 64, Adam optimizer, and an initial learning rate of 0.001 decreased by a factor of 4 every 20 epochs.
When using PN++ 
we considered a batch size of 64, SGD optimizer with momentum 0.9, and an initial learning rate of 0.01, decreased by a factor of 2 every 20 epochs.
Data augmentation is performed following verbatim the procedure proposed by \cite{pointnet_CVPR17}, \ie random vertex jittering drawn from $\mathcal{N}(0, 0.01)$, and random rotation around the shape elongation axis.

\vspace{-3mm}\paragraph{Baselines.} For all our experiments we use as reference the standard supervised baseline. 
It is a na\"ive variant of our method obtained by setting $\alpha=0$ which simply corresponds to turning off the puzzle solver.

\vspace{-3mm}\paragraph{Single Domain.}
We start by evaluating the performance of our approach in the most classical scenario when a single domain is available and present the results in Table~\ref{tab:shape_class}. We consider as testbed ModelNet40 ($C$=40) and ScanObjectNN ($C$=16) and observe that in both cases our multi-task approach consistently outperforms the baseline regardless of the used backbone.
These results indicate that by simply solving the auxiliary self-supervised task the learned representation is better able to capture the object semantics and provides further discriminative information to the final classifier. The advantage appears particularly evident on the real world dataset ScanObjectNN with a gain of about 3 percentage points (pp) for the difficult \texttt{PB\_T50\_RS\_BG} on PN.

\begin{table}[tb!]
    \centering
    \caption{Shape classification accuracy (\%) of our multi-task approach when the training set contains a limited amount of annotated data}
    \begin{tabular}{l@{~~~~}|@{~~~~}l@{~~~~}|@{~~~~}c@{~~~~}c@{~~~~}c@{~~~~}c}
        \hline
        \multirow{2}{*}{Backbone} & \multirow{2}{*}{Method} & \multicolumn{4}{c}{ModelNet40} \\
        & & 20\% & 40\% & 60\% & 100\% \\
        \hline
        \multirow{3}{*}{PN } & Baseline & 82.94 & 85.49 & 87.11 & 88.65 \\
            & Our FS & 82.09 & 87.03 & 88.13 & \textbf{89.71} \\
            & Our SS & \textbf{83.06} & \textbf{87.27} & \textbf{88.57} & - \\
        \hline
        \multirow{3}{*}{PN++ } & Baseline & 85.37 & 88.25 & 89.63 & 91.93 \\
            & Our FS & \textbf{86.35} & \textbf{89.59} & 89.18 & \textbf{92.10} \\
            & Our SS & 86.02 & 88.41 & \textbf{89.83} & - \\
        \hline
    \end{tabular}
    \label{tab:shape_class_limited_data}
    \vspace{-5mm}
\end{table}

\vspace{-3mm}\paragraph{Transfer Learning.} We also perform two TL experiments considering the availability of an extra unsupervised source $\mS'$. 
The first one considers $\mS$ = ModelNet40 and $\mS'$ = (5K samples from) ShapeNetCore and aims to analyze knowledge transfer among two different synthetic domains. The second one considers $\mS$ = ScanObjectNN and $\mS'$ = (4K samples from) ModelNet40 and targets knowledge transfer from synthetic to real point clouds. The cardinality of $\mS'$ was chosen to have a good balance between unsupervised ($\mS'$) and annotated ($\mS$) data.
Results are reported in the third and sixth row of Table~\ref{tab:shape_class}.
Overall we observe a further improvement up to 1 pp with respect to the previous results without transfer, with the only exception of ModelNet40 and \texttt{OBJ\_ONLY} when using the PN++ backbone. Here the extra synthetic information does not seem to provide useful information. Differently, for PN the advantage is always visible, indicating that also the backbone choice has a role in the transfer process.

We highlight that, although previous works discussed TL in combination with self-supervised tasks, their settings differs from ours.
In \cite{saudersievers} a two-stage pipeline is proposed: first a 3D puzzle solver is learned in an unsupervised manner on the whole ShapeNetCore dataset ($>$ 50K models), than the obtained weights are used to initialize a supervised model. On ModelNet40 this pipeline reaches an accuracy of 92.4\%, a negligible gain over the 92.2\% accuracy of the baseline with random initialization. Considering the amount of extra unsupervised data used and the different backbone (DGCNN~\cite{dgcnn}), our 92.1\%  obtained without extra information appears upstanding.
Another interesting comparison can be done with the recently proposed multi-task method~\cite{scanobjectnn_ICCV19} which combines classification and segmentation as a possible strategy to deal with real world point clouds. %
This approach obtains an accuracy of 80.2\% on \texttt{PB\_T50\_RS\_BG} over a baseline of 77.9\%, but requires an additional costly annotation phase  (foreground/background mask) being fully supervised. In contrast, our approach reaches 79.08\% accuracy without using any extra label, confirming the effectiveness of the auxiliary self-supervised task.

\begin{table*}[tb!]
    \centering 
    \caption{Shape classification accuracy (\%) of our multi-task when training and testing is done on different domains (DG). If the unlabeled target data is provided at training time (DA), our multi-task is able to adapt and reduce the domain gap}
    \resizebox{\textwidth}{!}{
    \begin{tabular}{l@{~~}l@{~}|@{~~}c@{~~}c@{~~}c@{~~}c@{~~}|@{~~}c}
    \hline
    \multicolumn{7}{c}{Domain Generalization and Adaptation} \\
    \hline 
    \multicolumn{2}{c@{~}|@{~~}}{\multirow{2}{*}{Method}} & 
    \multicolumn{4}{@{~~}c|@{~~}}{ModelNet40 $\rightarrow$} &
    \scriptsize{\texttt{PB\_T50\_RS\_BG}} $\rightarrow$ \\
    & &
    \scriptsize{\texttt{OBJ\_ONLY}} &
    \scriptsize{\texttt{OBJ\_BG}} & 
    \scriptsize{\texttt{PB\_T50\_RS}} &
    \scriptsize{\texttt{PB\_T50\_RS\_BG}} &
    ModelNet40 \\
    \hline
    \multicolumn{2}{l@{~~}|@{~~}}{PointDAN~\cite{pointdan10}}
         & 56.42 & 44.84 & 48.99 & 34.39 & 54.66 \\
    \hline
        \multirow{3}{*}{PN}
        & Baseline & 54.74 & 43.58 & 44.96 & 34.25 & 47.43 \\
        & Our DG & 54.53 & 49.68 & 45.22 & 36.28 & 39.30 \\
        & Our DA & 58.53 & 47.58 & 46.70 & 35.85 & 51.54 \\
    \hline
        \multirow{3}{*}{PN++}
        & Baseline & 52.49 & 44.00 & 44.83 & 34.29 & 47.66 \\
        & Our DG & 57.47 & 52.42 & 52.84 & 38.65 & 52.88 \\
        & Our DA & \textbf{60.4} & \textbf{53.89} & \textbf{54.66} & \textbf{39.63} & \textbf{56.07} \\
    \hline\hline
        \multicolumn{2}{l@{~}|@{~~}}{{3DmFV} \cite{ben20183dmfv}} & 30.90 & 24.00 & 24.90 & 16.40 & 51.50 \\
        \multicolumn{2}{l@{~}|@{~~}}{{PointCNN} \cite{NIPS2018_7362}} & 32.20 & 29.50 & 24.60 & 19.20 & 49.20 \\
        \multicolumn{2}{l@{~}|@{~~}}{{PointNet} \cite{pointnet_CVPR17}} & 42.30 & 41.10 & 31.10 & 23.20 & 50.90 \\
        \multicolumn{2}{l@{~}|@{~~}}{{PointNet++} \cite{pointnet++_NIPS2017}~~~} & 43.60 & 37.70 & 32.00 & 22.90 & 47.40 \\
        \multicolumn{2}{l@{~}|@{~~}}{{SpiderCNN} \cite{xu2018spidercnn}} & 44.20 & 42.10 & 30.90 & 22.20 & 46.60 \\
        \multicolumn{2}{l@{~}|@{~~}}{{DGCNN}\cite{dgcnn}} & 49.30 & 46.70 & 36.80 & 27.20 & 54.70 \\
    \hline    
    \end{tabular}}
    \label{tab:DGDA}
    \vspace{-3mm}
\end{table*}

\vspace{-3mm}\paragraph{Few-Shot and Semi-Supervised.}
We focused on ModelNet40 for experiments in these settings. The results in Table~\ref{tab:shape_class_limited_data} show the performance obtained when the amount of labeled training data reduces up to only 20\% of the original amount. We can observe that, despite the overall drop in performance, our multi-task approach in the few-shot (FS) setting maintains its advantage with respect to the baseline. When considering also the unlabeled data for the semi-supervised (SS) setting and the PN backbone, we observe a further increase in performance.

\vspace{-3mm}\paragraph{Domain Generalization.}
When training and test data are drawn from two very different distributions the model learned on the former ones usually fails to generalize to the latter. Being able to maintain a good performance in this challenging condition is crucial in all the cases in which obtaining annotated data of the target domain is not possible.
We consider the DG setting when training on ModelNet40 and testing on ScanObjectNN and report results in Table~\ref{tab:DGDA}. Our multi-task approach fully learned only on synthetic data shows a significant improvement with respect to the baseline with gains up to 6 and 8 pp in the \texttt{OBJ\_BG} and with a still relevant gain of 2 and 4 pp in the most challenging \texttt{PB\_T50\_RS\_BG}, respectively with PN and PN++.
We also consider the inverse generalization direction from \texttt{PB\_T50\_RS\_BG} to ModelNet40. Here the training data is affected by various sources of noise and adding the self-supervised task on PN seems to backfire, increasing the risk of overfitting. On the other way round, PN++ is more reliable and presents a significant gain of 4 pp.

\begin{figure*}[!t]
\centering
\begin{tabular}{cc}
\includegraphics[width=0.6\textwidth]{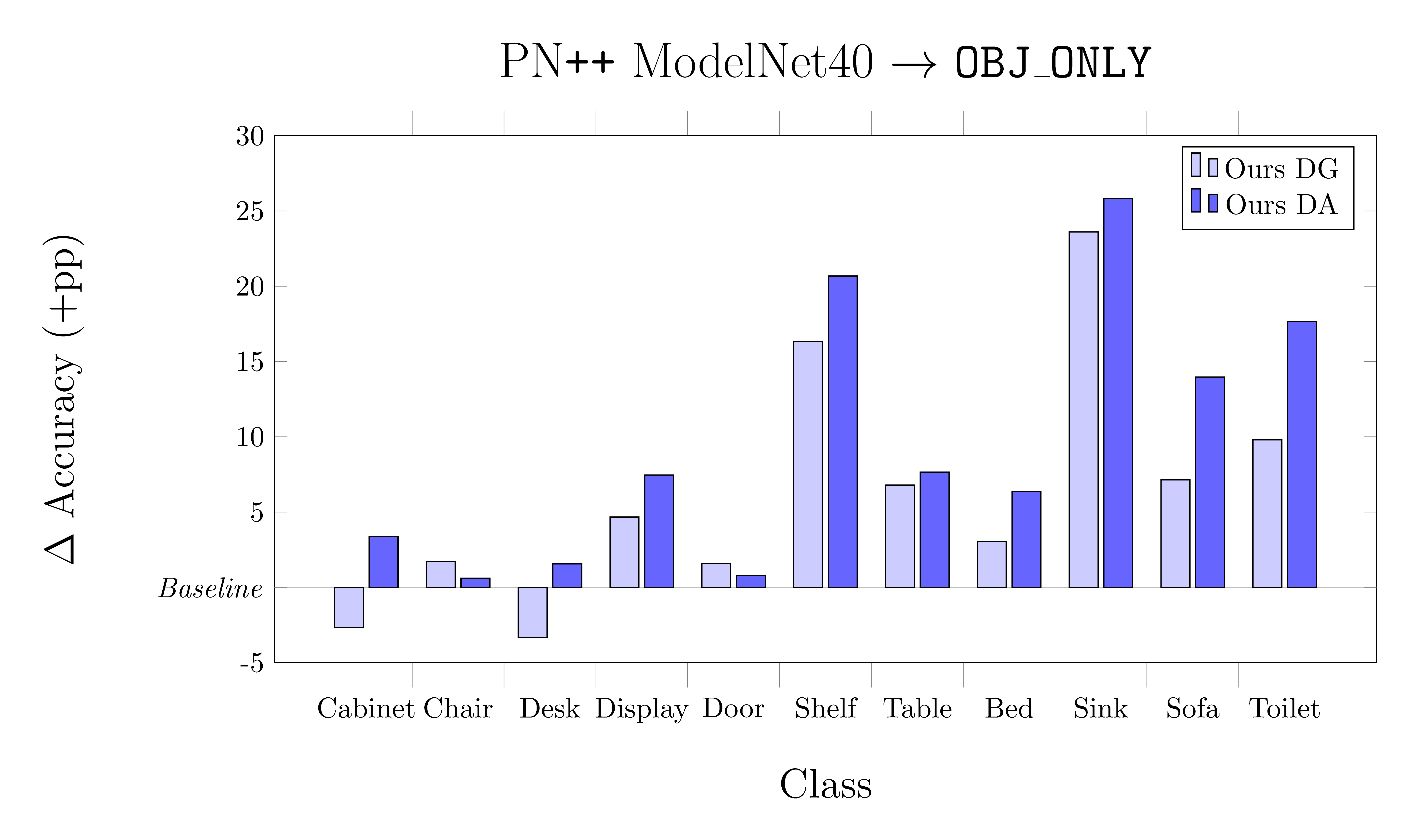} &
\includegraphics[width=0.4\textwidth]{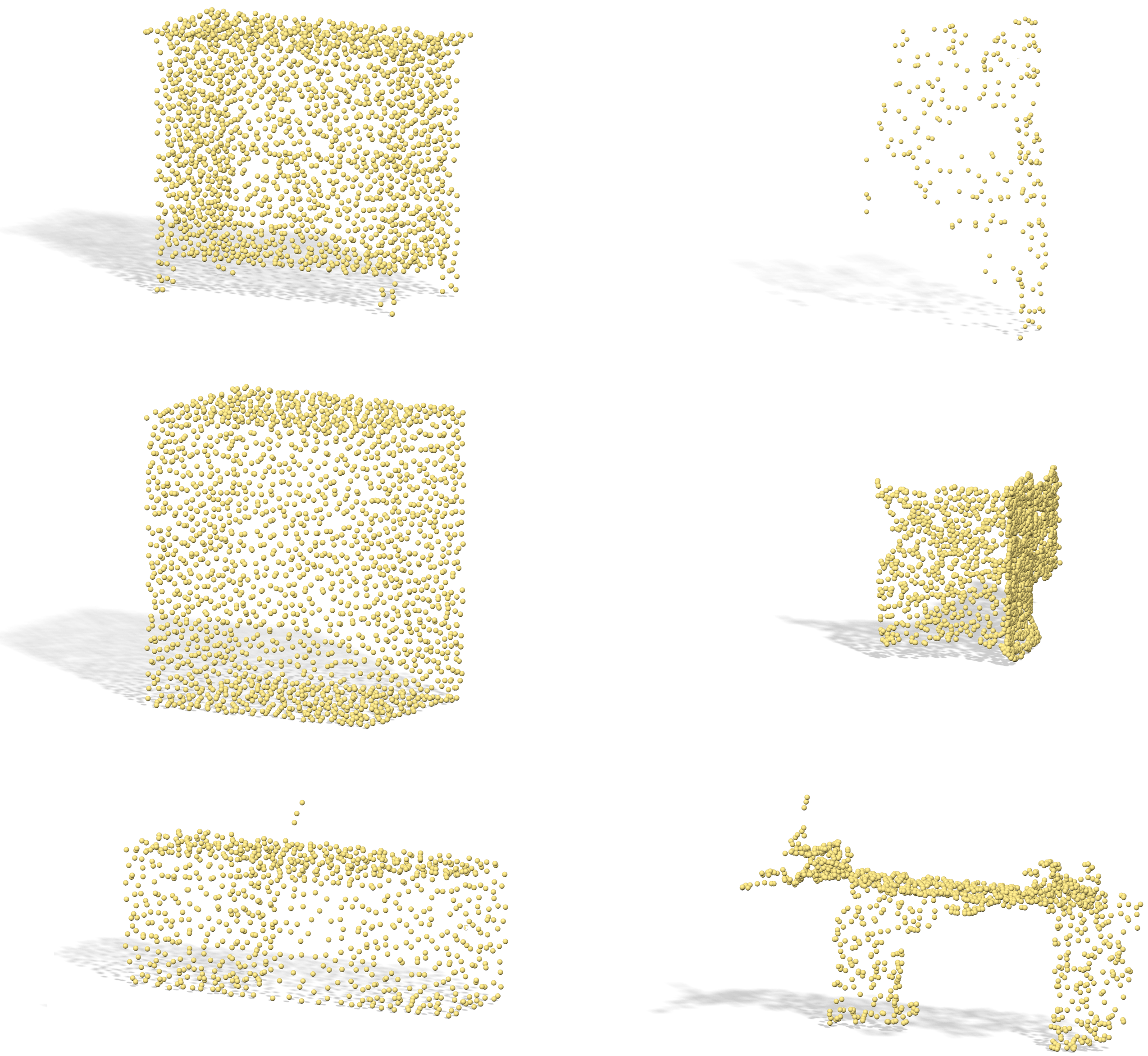}
\end{tabular}
\caption{\emph{left}: Per-class accuracy improvements (w.r.t. baseline)
: our approach tackles domain shift with a gain up to 25 pp.
\emph{right}: Real-World shapes from classes `Cabinet' and `Desk' are very difficult to classify across domains due to noise and occlusion. Here the first column show examples from ModelNet, while the second column from ScanObject. First two rows are cabinet, the last row is desk}
\vspace{-5mm}
\label{fig:delta_perclass_accuracy}
\end{figure*}

\paragraph{Domain Adaptation.} We also investigated whether our multi-task approach could close the domain gap when unlabeled target data are available at training time.
DA results in Table~\ref{tab:DGDA} provide a positive answer showing a further increase in performance over the DG results (with few exceptions with PN). A detailed per-class analysis of both DA and DG results with the PN++ backbone for the ModelNet40$\to$\texttt{OBJ\_ONLY} case is shown in Figure~\ref{fig:delta_perclass_accuracy}. This visualization helps to understand which are the most difficult cases across domains.

The very recent PointDAN method \cite{pointdan10} proposed to solve point-cloud domain shifts by combining local and global alignment. Local alignment is obtained trough an attention module that takes into consideration the relationship between close nodes, while global alignment is performed through maximum classifier discrepancy \cite{saito2017maximum}. Table~\ref{tab:DGDA} shows that our multi-task approach largely outperforms this solution\footnote{See the Appendix for an even more extensive comparison.}. 
Finally, an overall look 
at the performance of several recent point cloud networks is provided in the bottom part of Table~\ref{tab:DGDA}.
The results indicate that our multi-task approach establishes the new state-of-the-art for classification on real world data from synthetic training. Even in the opposite learning direction from real to synthetic, our model combining supervised and self-supervised learning shows promising results: the ability to adapt provides it with a further way to improve over existing references.

\subsection{Part Segmentation Results}
The second set of our experiments is dedicated to part segmentation, the problem of assigning each vertex to the shape part to which it belongs to. 
By following \cite{pointnet_CVPR17}, the quality of the predicted part segmentation is evaluated in terms of the mean Intersection-over-Union (mIoU) metric. The mIoU of a shape is defined as the average over its $Q$ parts of the IoU between the ground-truth and predicted segmentations of each part. The mIoU of a category is defined as the average of the mIoUs of the shapes it contains.

\vspace{-3mm}\paragraph{Training Details.}
Throughout all the part segmentation experiments we used the PointNet Segmentation backbone from \cite{pointnet_CVPR17}. We slightly modified the network architecture introducing a branch for jointly solving the 3D puzzle task, this branch shares most of the initial network layers with the main segmentation one. Our modifications does not increase the original segmentation branch capacity. We used batch size of 64, Adam optimizer, and an initial learning of rate 0.001, decreased by a factor of 2 every 20 epochs. Data augmentation is applied exactly as in our classification experiments.

\vspace{-3mm}\paragraph{Single Domain and Transfer Learning.}
Table~\ref{tab:scanobjectchair} shows the part segmentation results obtained by our method on the chair shapes from two challenging subsets of ScanObjectNN in terms of the evaluation metric used in \cite{scanobjectnn_ICCV19}.
In the SD setting the introduction of self-supervision in the learning process does not improve over the baseline accuracy.
In the TL setting, by considering as extra source of knowledge $\mS'$ the unlabeled chairs from ModelNet40, our approach proves once again its effectiveness.

\begin{figure*}[tb!]
\centering
\begin{overpic}
    [width=\linewidth]{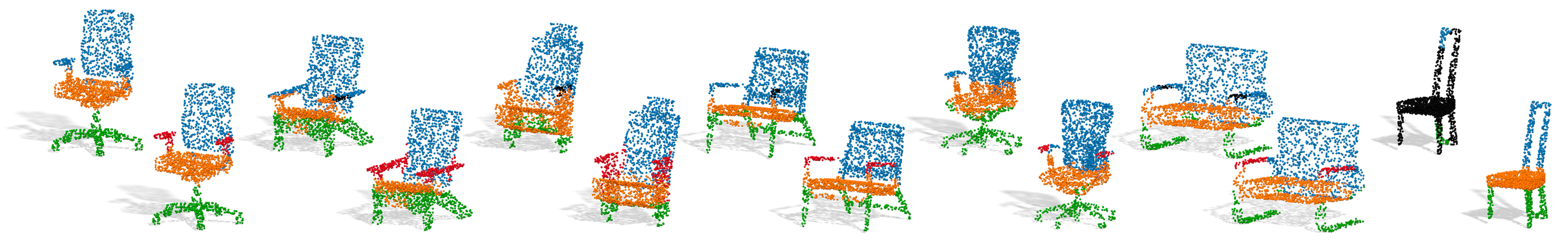}
\end{overpic}
\begin{overpic}
    [width=\linewidth]{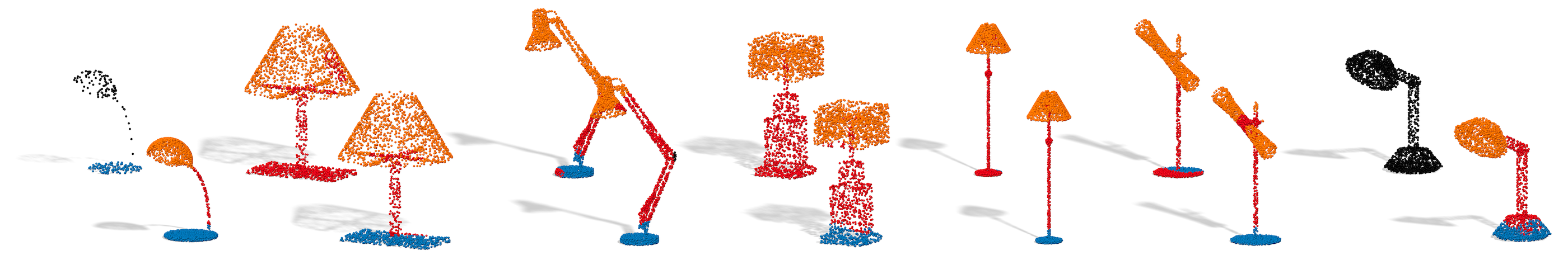}
\end{overpic}
\caption{Part segmentation of chairs and lamps when only 1\% of training data are available. Each couple shows the baseline prediction (top left) and our approach (bottom right). The last example show the worst case for the baseline. Black points denotes predictions whose maximum value was not a chair or lamp part}
\label{fig:part_segm_ss}
\vspace{-3mm}
\end{figure*}

\vspace{-3mm}\paragraph{Few-Shot and Semi-Supervised.}
By following \cite{Hassani_2019_ICCV} we randomly sample 1\% and 5\% of the ShapeNetPart train set to evaluate the point features in a semi-supervised setting. The results in Table \ref{tab:part_segm_iccv19} indicate that our multi-task approach, although not improving over the baseline in the few-shot setting, in the semi-supervised setting outperforms the current state of the art in the 1\% case and practically matches it in the 5\% case.  It is interesting to underline that also our best competitor CCD \cite{Hassani_2019_ICCV} is a multi-task approach that combines clustering and reconstruction with a self-supervised classification obtained by learning on the clustering auto-defined labels.
For a more in-depth analysis of our results we plot some visualizations out of our 1\% part segmentation experiment in Figure \ref{fig:part_segm_ss} for chairs and lamps. Regarding chairs, our multi-task approach seems to allow a better recognition of the armrests. Indeed the position of these relative small parts of the chair may be better learned thanks to the puzzle solution task. A similar consideration may be done for the lamp basis.

\begin{table*}[tb!]
    \centering
    \caption{Part segmentation of chairs from two variants of ScanObjectNN evaluated in terms of per part average accuaracy (\%) and overall accuracy (\%)}
    \resizebox{\textwidth}{!}{
    \begin{tabular}{l@{~~}|@{~}l@{~}|@{~~}c@{~~}c@{~~}c@{~~}c@{~~}c@{~~}|@{~~}cc}
        \hline
            Dataset & Method & Bg & Seat & Back & Base & Arm & Avg. Acc. & Overall Acc. \\
        \hline
            \multirow{3}{*}{\scriptsize{\texttt{OBJ\_BG}}}
            & Baseline & 65.14 & \textbf{87.88} & \textbf{89.73} & 67.16 & 58.97 & 73.02 & 81.62 \\
            & Our SD & 64.97 & 87.46 & 86.27 & 68.96 & 57.54 & 73.04 & 81.67 \\
            & Our TL & \textbf{69.43} & 86.59 & 88.71 & \textbf{72.70} & \textbf{61.37} & \textbf{75.76} & \textbf{82.60} \\ 
        \hline
            \multirow{3}{*}{\scriptsize{\texttt{PB\_T50\_R\_BG}}}
            & Baseline & 82.06 & \textbf{83.71} & 75.30 & \textbf{54.75} & \textbf{35.11} & \textbf{66.19} & 81.13 \\
            & Our SD & 82.02 & 83.50 & 77.80 & 51.53 & 25.50 & 64.07 & 81.31 \\
            & Our TL & \textbf{83.08} & 81.87 & \textbf{79.06} & 50.71 & 30.40 & 64.99 & \textbf{81.82} \\
        \hline
    \end{tabular}
    }
    \label{tab:scanobjectchair}
    \vspace{-2mm}
\end{table*}
\begin{figure}[tb!]
\begin{minipage}[t]{0.4\textwidth}
    \centering
    \small
    \captionsetup{width=\textwidth}
    \captionof{table}{Accuracy (mIoU) for part segmentation on ShapeNetPart with limited annotations.}
    \begin{tabular}{lcc}
        \hline
        Method & 1\% & 5\% \\
        \hline
        SO-Net \cite{sonet_2018_CVPR} & 64.00 & 69.00 \\ 
        PointCapsNet \cite{Zhao_2019_CVPR} & 67.00 & 70.00 \\
        CCD \cite{Hassani_2019_ICCV} & 68.20 & \textbf{77.70} \\
        \hline
        Baseline      & 64.52 & 75.75 \\
        Our FS  & 64.49 & 75.07 \\
        Our SS  & \textbf{71.95} & 77.42\\
        \hline
    \end{tabular}
    \label{tab:part_segm_iccv19}
\end{minipage}
\hfill
\begin{minipage}[t]{0.6\textwidth}
    \vspace{3mm}
    \centering
    \small
    \captionsetup{width=.9\textwidth}
    \captionof{table}{Per part and average accuracy (\%) of chair segmentation. We use the same metric of Table \ref{tab:scanobjectchair}}
    \begin{tabular}{l|cccc|c@{}}
    \hline 
    \multicolumn{6}{c}{Part Segmentation - DA/DG} \\
        \hline
            \multicolumn{6}{c}{\footnotesize{ShapeNetPart $\rightarrow$ \texttt{OBJ\_BG}}} \\
        \hline
            \footnotesize{Method} & \footnotesize{Seat} & \footnotesize{Back} & \footnotesize{Base} & \footnotesize{Arm} & \footnotesize{Avg.} \\
        \hline
            \footnotesize{Baseline} & 67.85 & 45.60 & 84.89 & 14.87 & 53.30  \\
             \footnotesize{our DG} & \textbf{71.80} & 42.61 & 84.57 & \textbf{21.48} & 55.11  \\
            \footnotesize{our DA} & 65.70 & \textbf{49.11} & \textbf{85.91} & 21.40 & \textbf{55.53}  \\
        \hline
    \end{tabular}
    \label{tab:scanobjectchairDADG}
\end{minipage}
\vspace{-2mm}
\end{figure}

\vspace{-3mm}\paragraph{Domain Generalization and Adaptation.} We focus on the domain shift between synthetic and real chairs with ShapeNetPart as source and ScanObjectNN as target.
We highlight that ScanObjectNN has some part annotation issue confirmed by the authors through personal communications, thus we prefer to use only the \texttt{OBJ\_BG} provided subdomain, neglecting the background which is absent in ShapenNetPart.
Table \ref{tab:scanobjectchairDADG} collects the results confirming also in this case the advantage of our multi-task approach over the supervised learning baseline.

\subsection{Ablation Analysis}
\label{subsection:ablation}
As indicated in sec. \ref{sec:method} our approach has two main hyperparameters.
One related to the learning model ($\alpha$) and the other needed to define the puzzled data ($l$). We analyze how much our method is sensitive to their variation by considering different values of $\alpha=\{0.4, 0.6, 0.8\}$ and three different puzzle decomposition settings with $l=\{2,3,4\}$. In particular we focus on the DA shape classification with PN backbone on ModelNet40 as source and $\texttt{OBJ\_ONLY}$, $\texttt{PB\_T50\_RS\_BG}$ as target. We also consider part segmentation SS with PN segmentation backbone on 1\% of ShapeNetPart. 
\begin{figure}[tb]
    \centering
    \begin{tabular}{ccc}
    \hspace{-2mm}\includegraphics[height=0.25\textwidth]{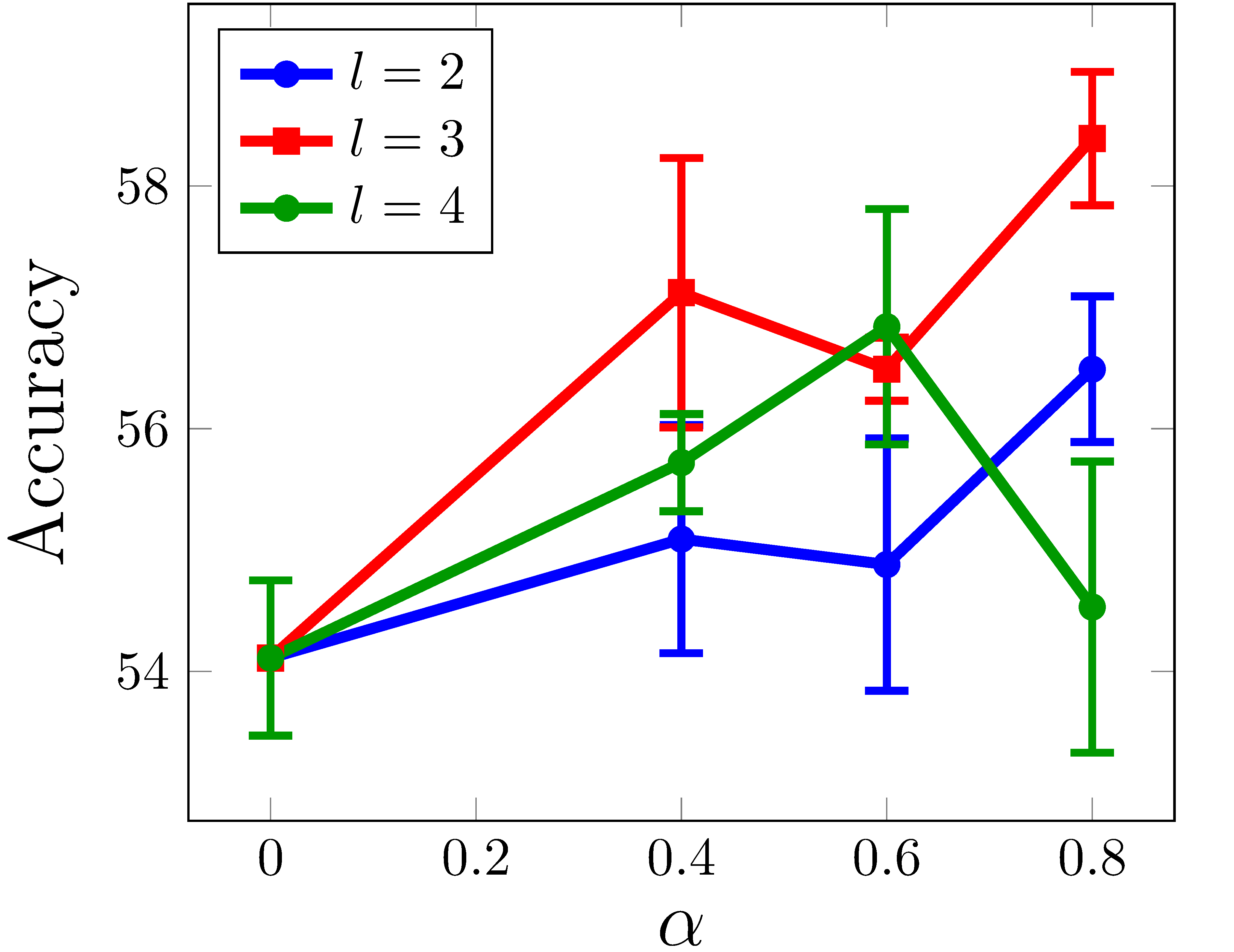} & 
    \hspace{-4mm}\includegraphics[height=0.25\textwidth]{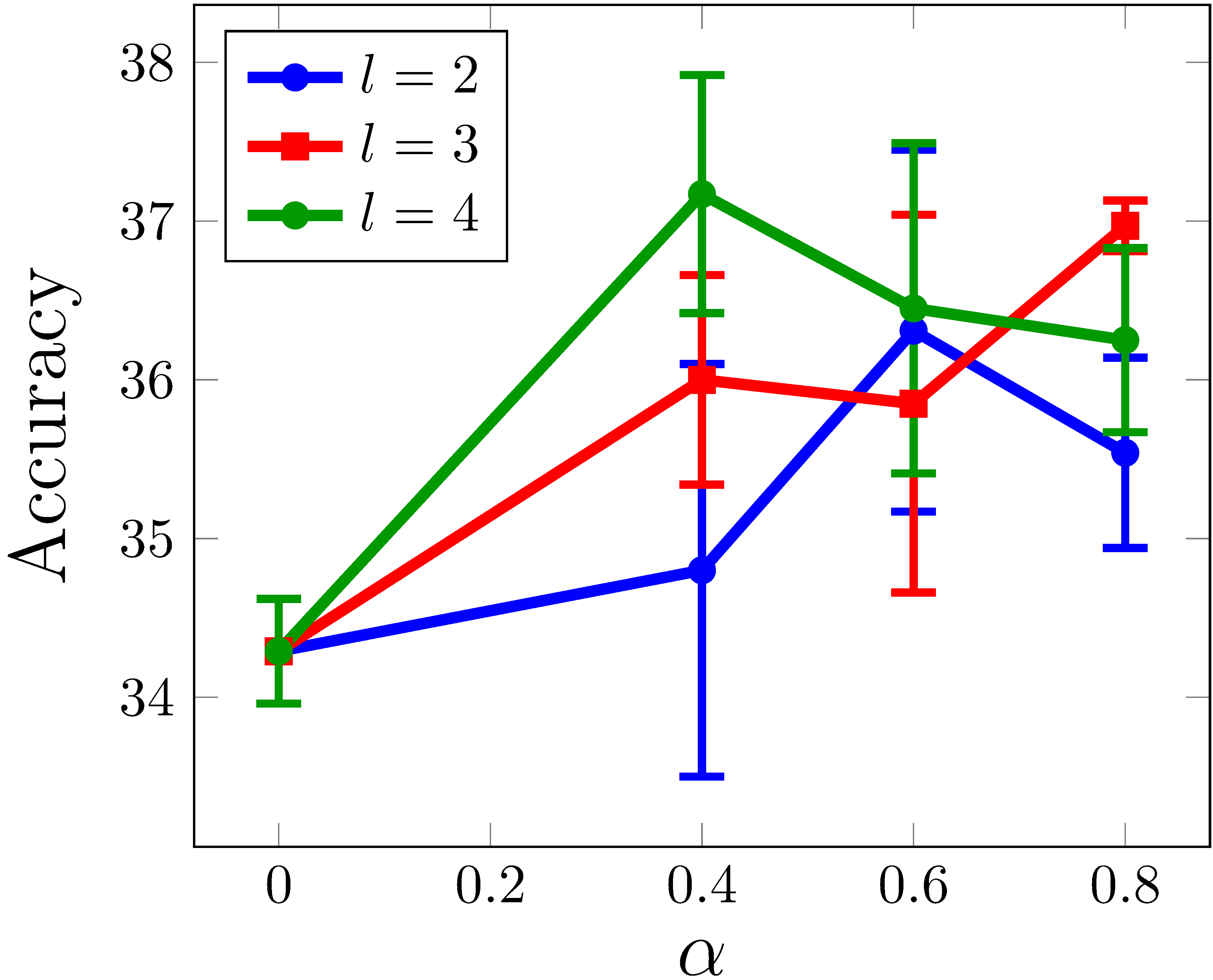} &
    \hspace{-4mm}\includegraphics[height=0.25\textwidth]{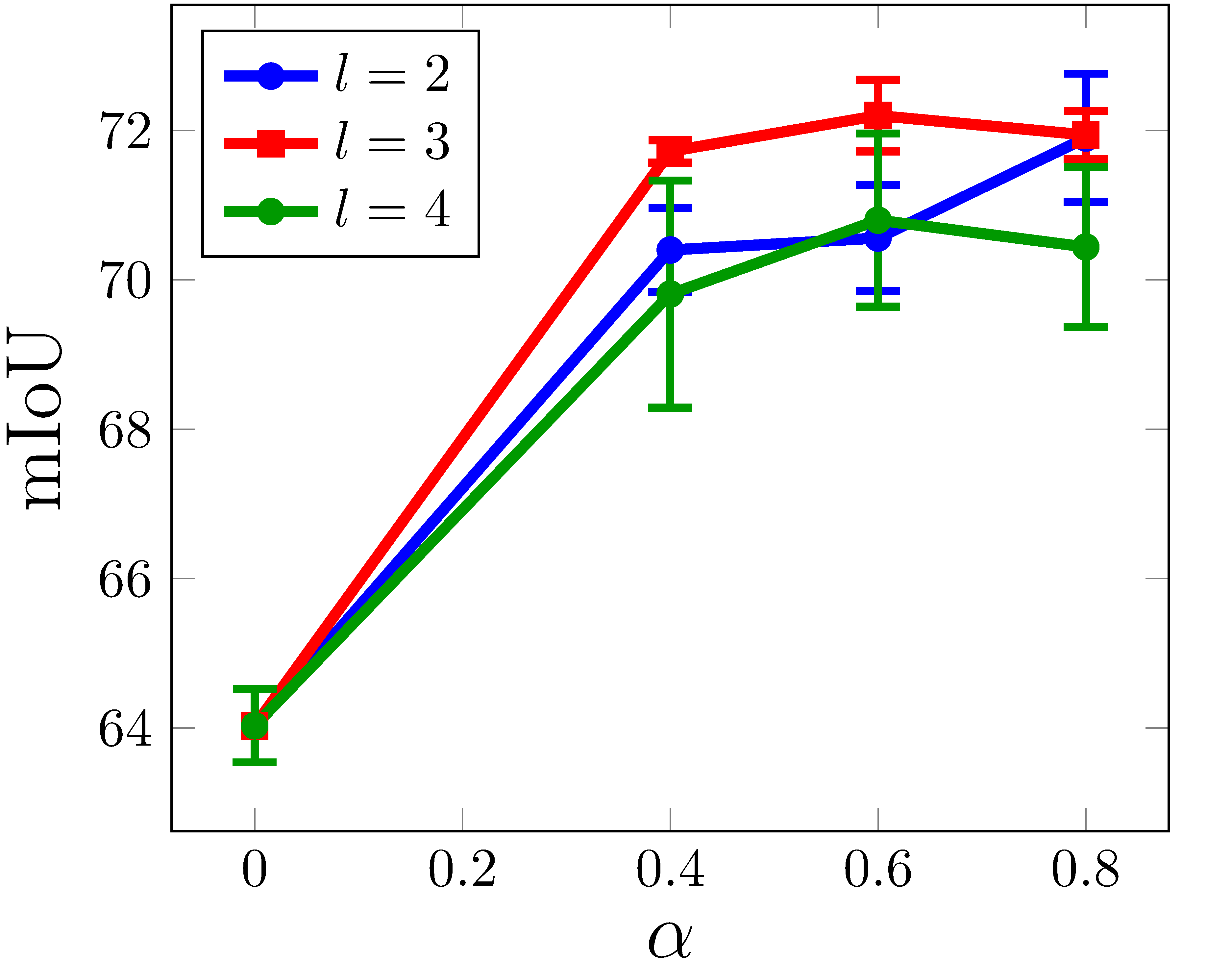}\\
    \vspace{-2mm}
    \scriptsize{(a) DA ModelNet40$\rightarrow$\texttt{OBJ\_ONLY}} & 
    \hspace{4mm}\scriptsize{(b) DA ModelNet40$\rightarrow$\texttt{PB\_T50\_RS\_BG}} & 
    \scriptsize{(c) SS ShapeNetPart 1\%} \\ 
    \end{tabular}
    \caption{Parameter ($\alpha$, $l$) evaluation in case of cross domain Shape Classification (a,b), and  1\% semi-supervised Part Segmentation (c). The case $\alpha=0$ corresponds to the baseline regardless of the $l$ value. Each experiment is repeated three times and we report here the average results with their standard deviation}
    \label{fig:ablation_plot}
    \vspace{-2mm}
\end{figure}
Figure \ref{fig:ablation_plot} confirms that on average $l=3$ is the best choice: intermediate between the minimum decomposition of an object into $2^3=8$ big parts and the very fine decomposition into $4^3=64$ parts. On the other hand, for the puzzle loss weight, our standard choice $\alpha=0.6$ can be improved by passing to $\alpha=0.8$, which indicates that it is possible to get an even further advantage with an ad hoc finetuned choice of the parameters. Overall there is a trade off between the two considered hyper-parameters: the auxiliary task can have a low weight as far as the number of puzzle part is high and vice-versa. 
This discussion holds both for DA shape classification and part segmentation results. In the latter case the difficulty of dealing with $l=4$ is even more evident.

\subsection{Qualitative Analysis of Puzzle Solution} \label{subsection:qualitative}
\begin{figure*}[tb!]
\begin{minipage}{0.195\linewidth}
\centering
\begin{overpic}[width=\linewidth, clip, trim=250 0 400 0]
    {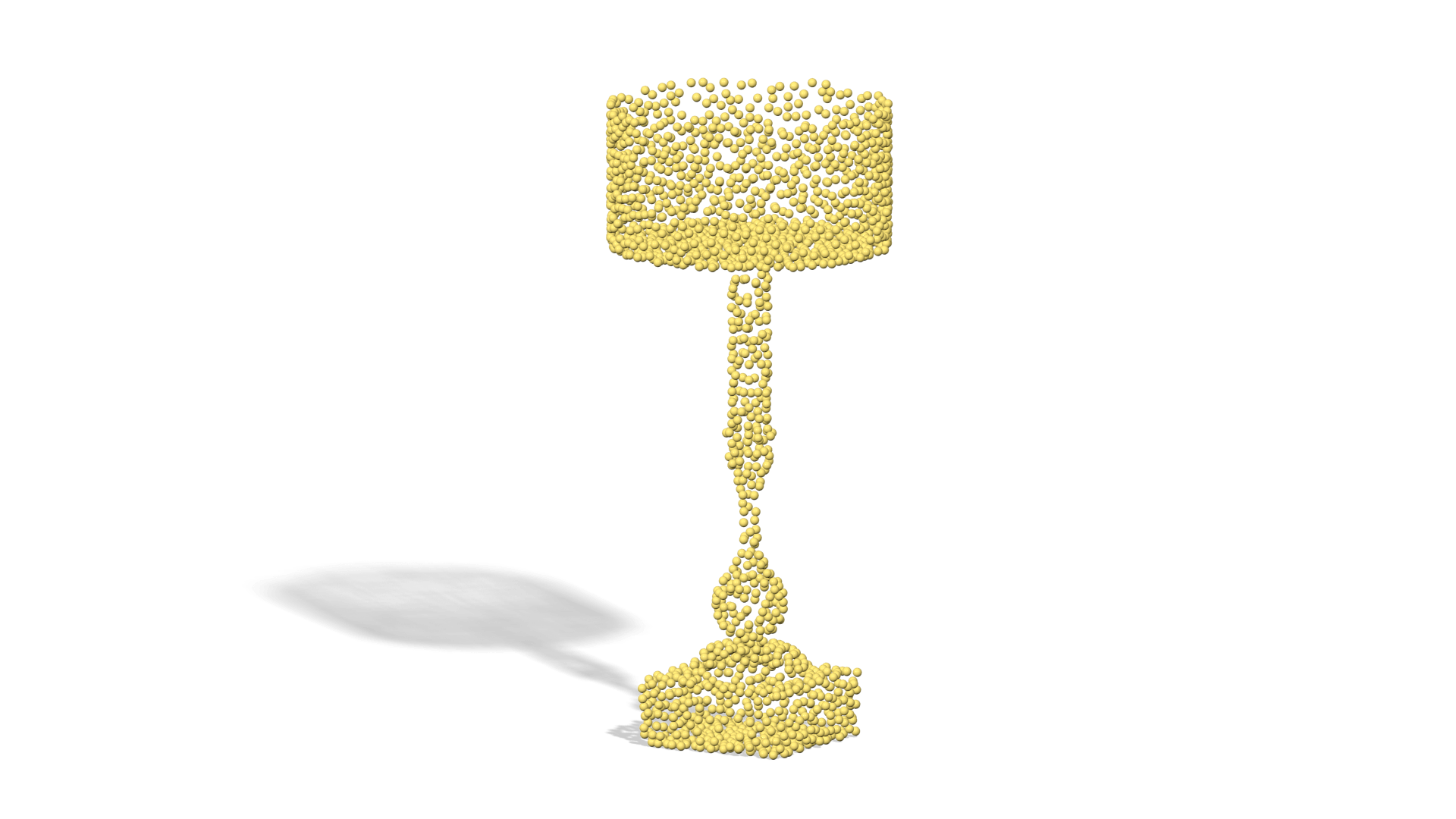}
\end{overpic}
\end{minipage}
\begin{minipage}{0.195\linewidth}
\centering
\begin{overpic}[width=\linewidth, clip, trim=250 0 250 0]
    {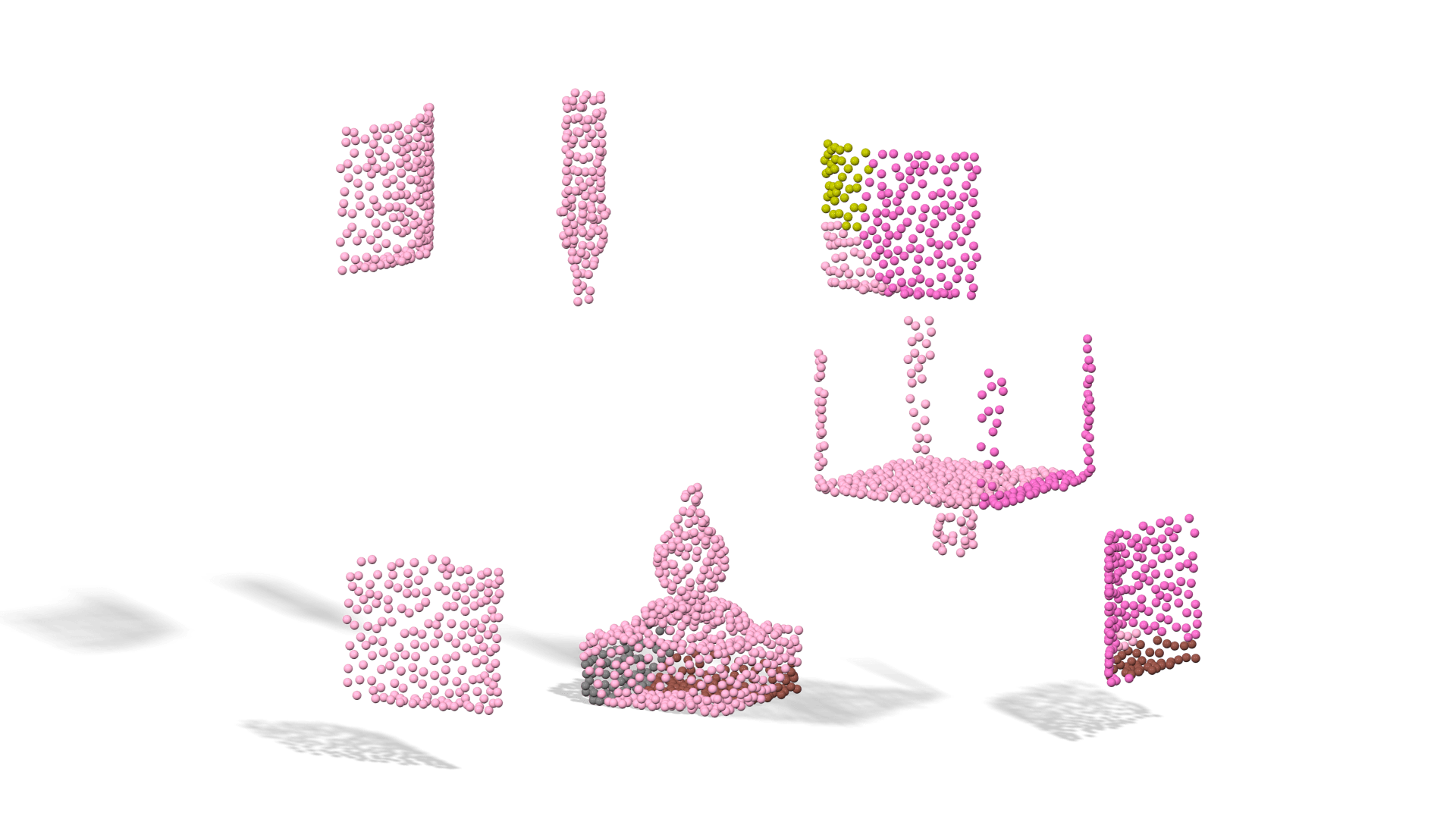}
\end{overpic}
\end{minipage}
\begin{minipage}{0.195\linewidth}
\centering
\begin{overpic}[width=\linewidth, clip, trim=250 0 250 0]
    {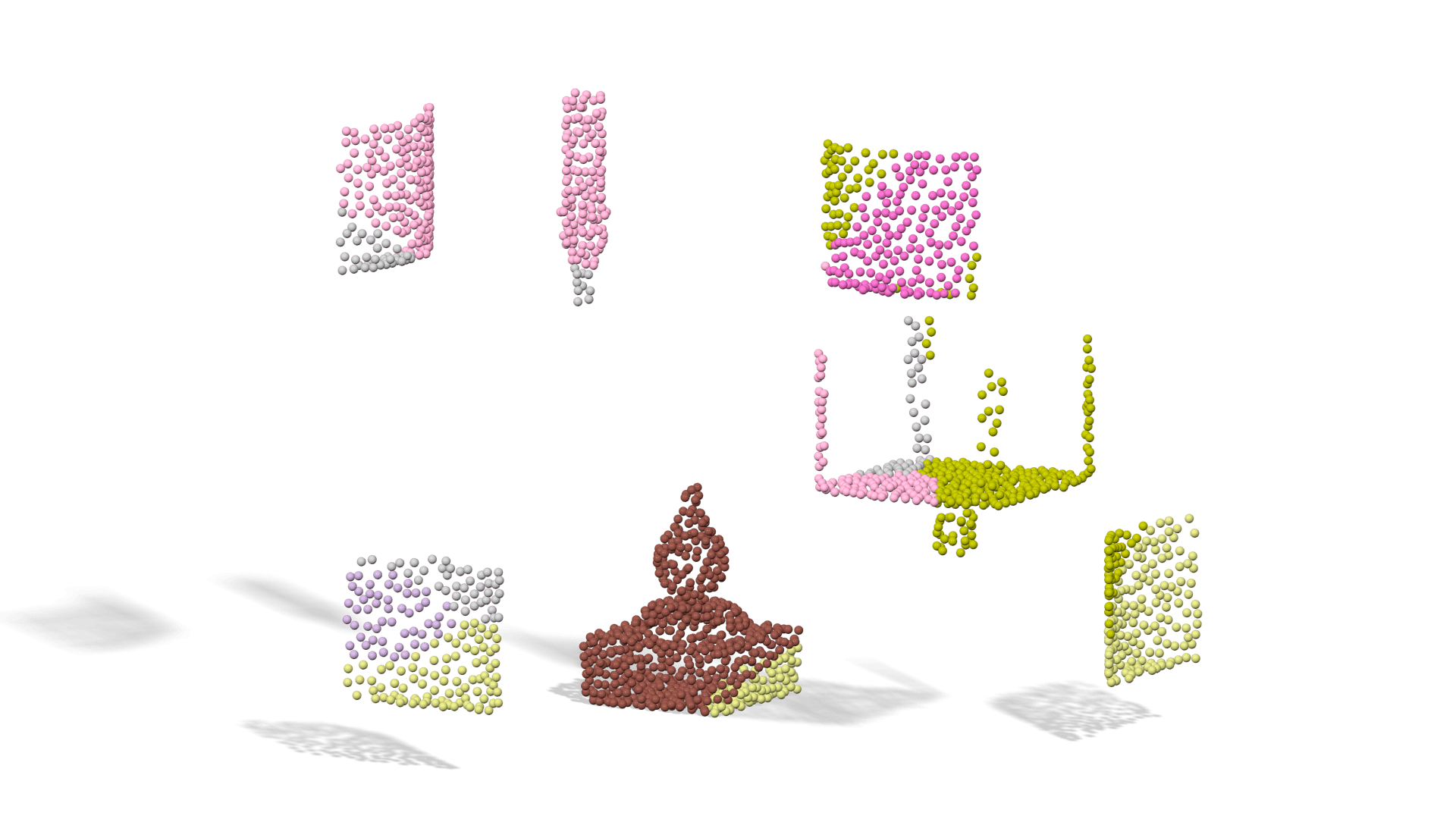}
\end{overpic}
\end{minipage}
\begin{minipage}{0.195\linewidth}
\centering
\begin{overpic}[width=\linewidth, clip, trim=250 0 250 0]
    {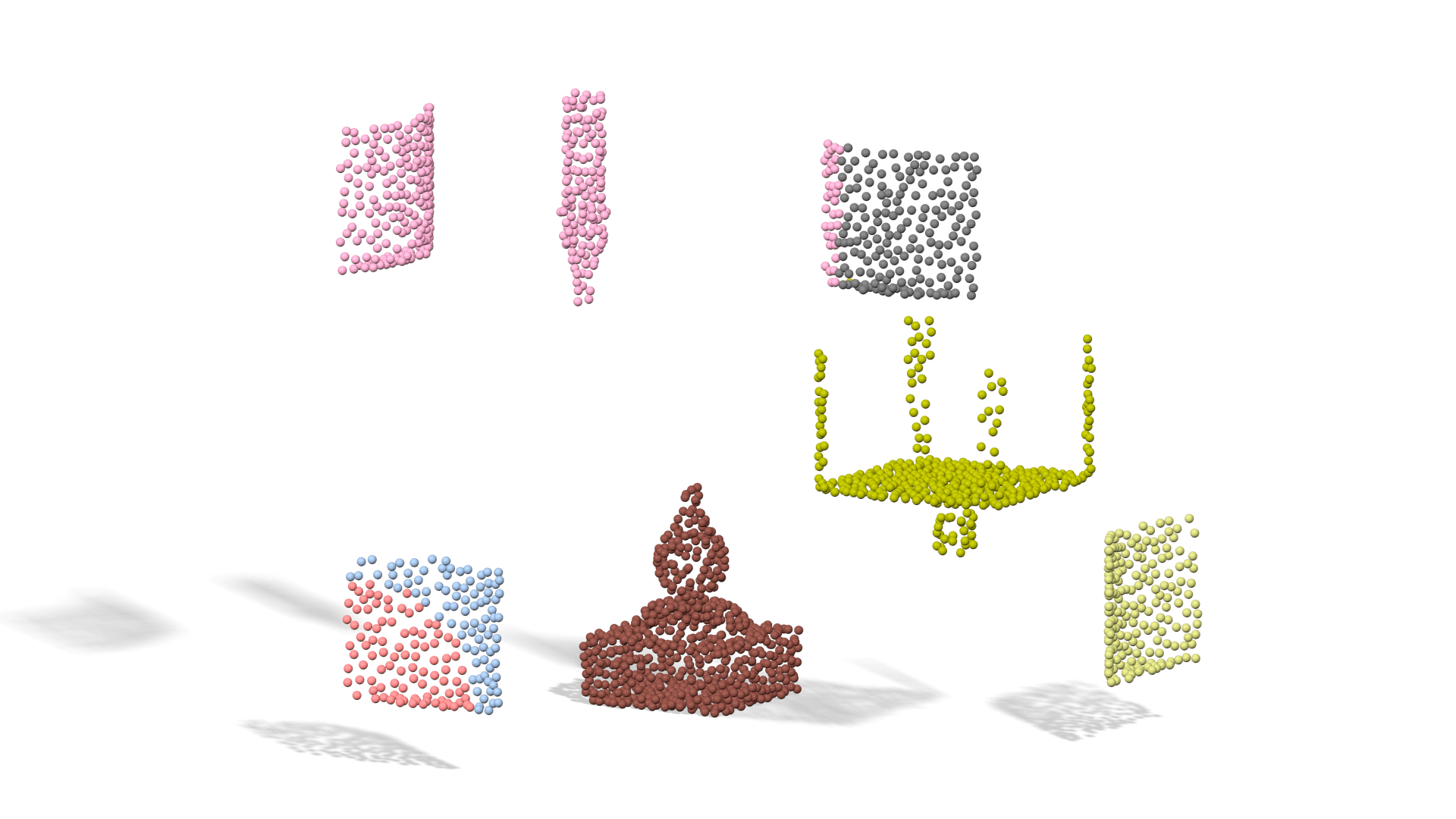}
\end{overpic}
\end{minipage}
\begin{minipage}{0.195\linewidth}
\centering
\begin{overpic}[width=\linewidth, clip, trim=250 0 250 0]
    {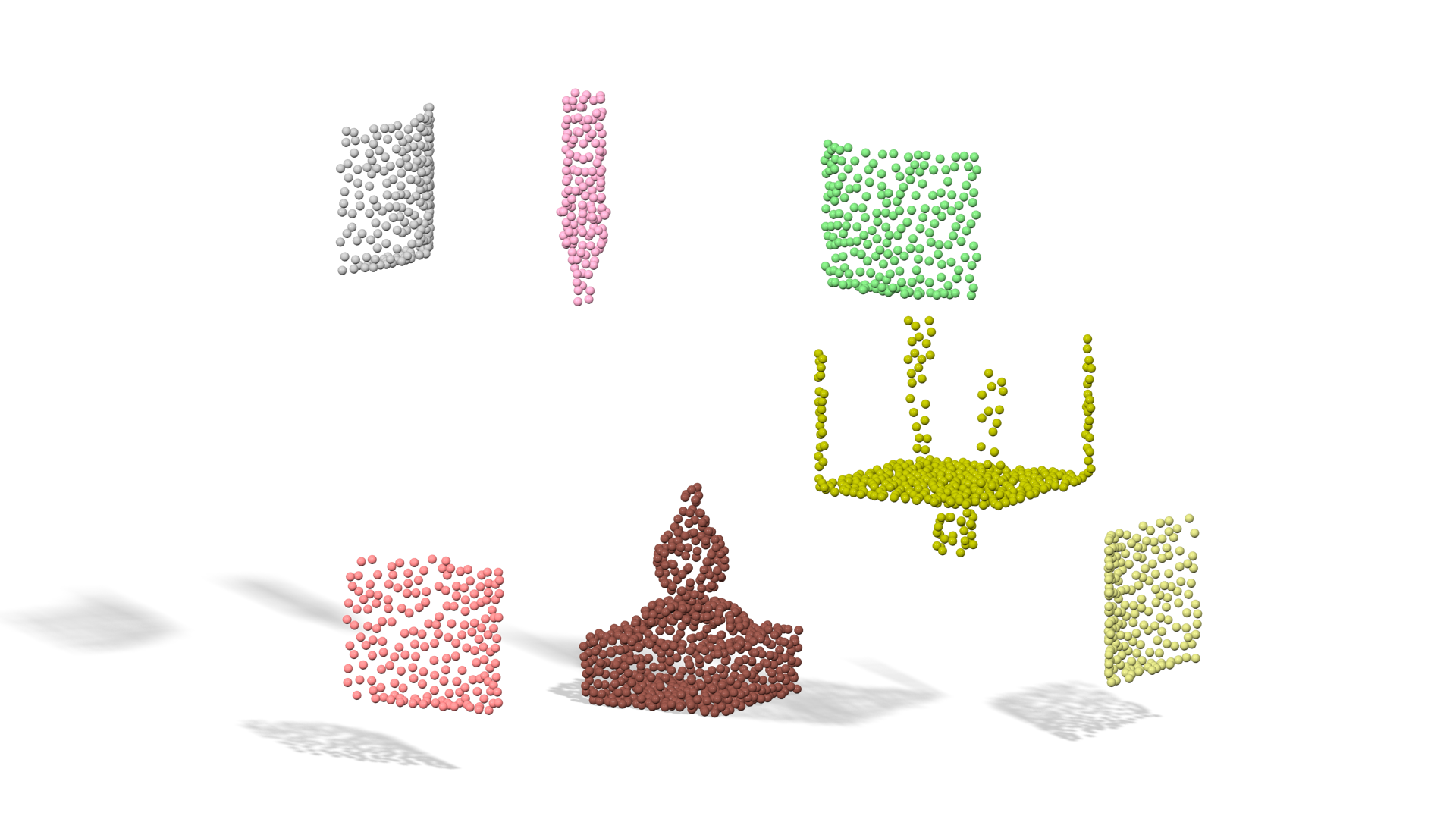}
\end{overpic}
\end{minipage} \\

\begin{minipage}{0.195\linewidth}
\centering
\begin{overpic}[width=\linewidth, clip, trim=250 0 400 0]
    {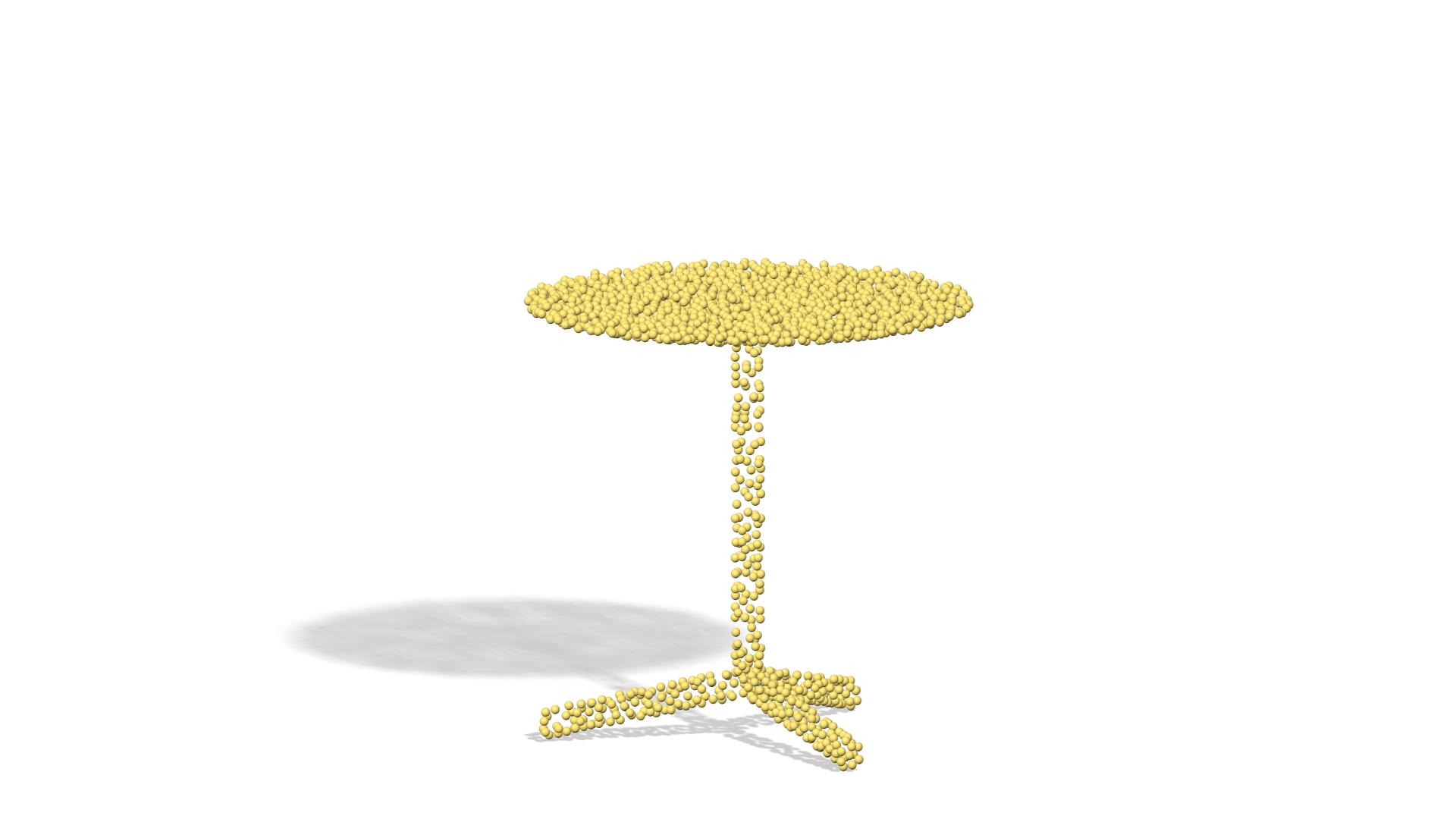}
\end{overpic}
\end{minipage}
\begin{minipage}{0.195\linewidth}
\centering
\begin{overpic}[width=\linewidth, clip, trim=250 0 400 0]
    {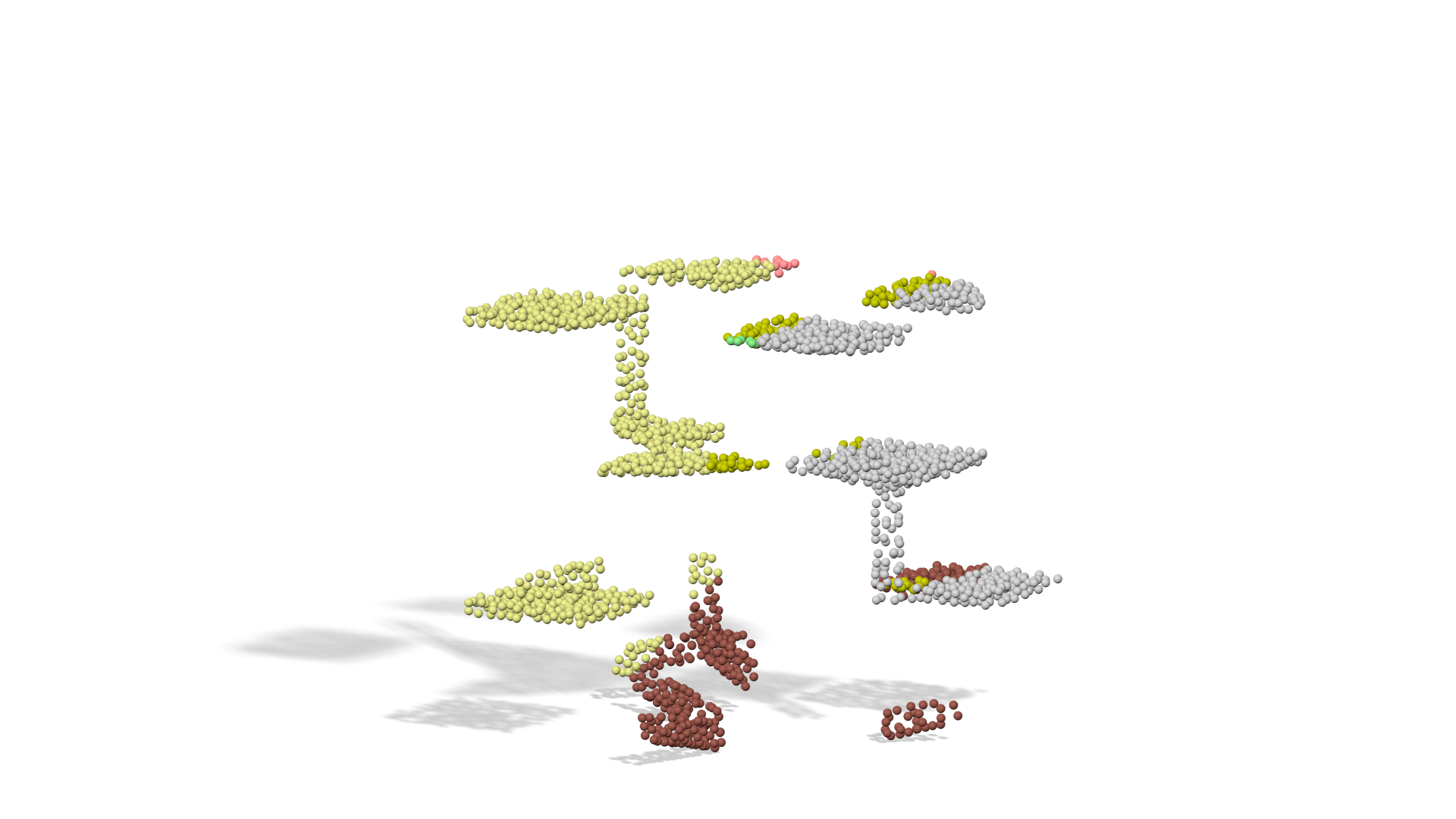}
\end{overpic}
\end{minipage}
\begin{minipage}{0.195\linewidth}
\centering
\begin{overpic}[width=\linewidth, clip, trim=250 0 400 0]
    {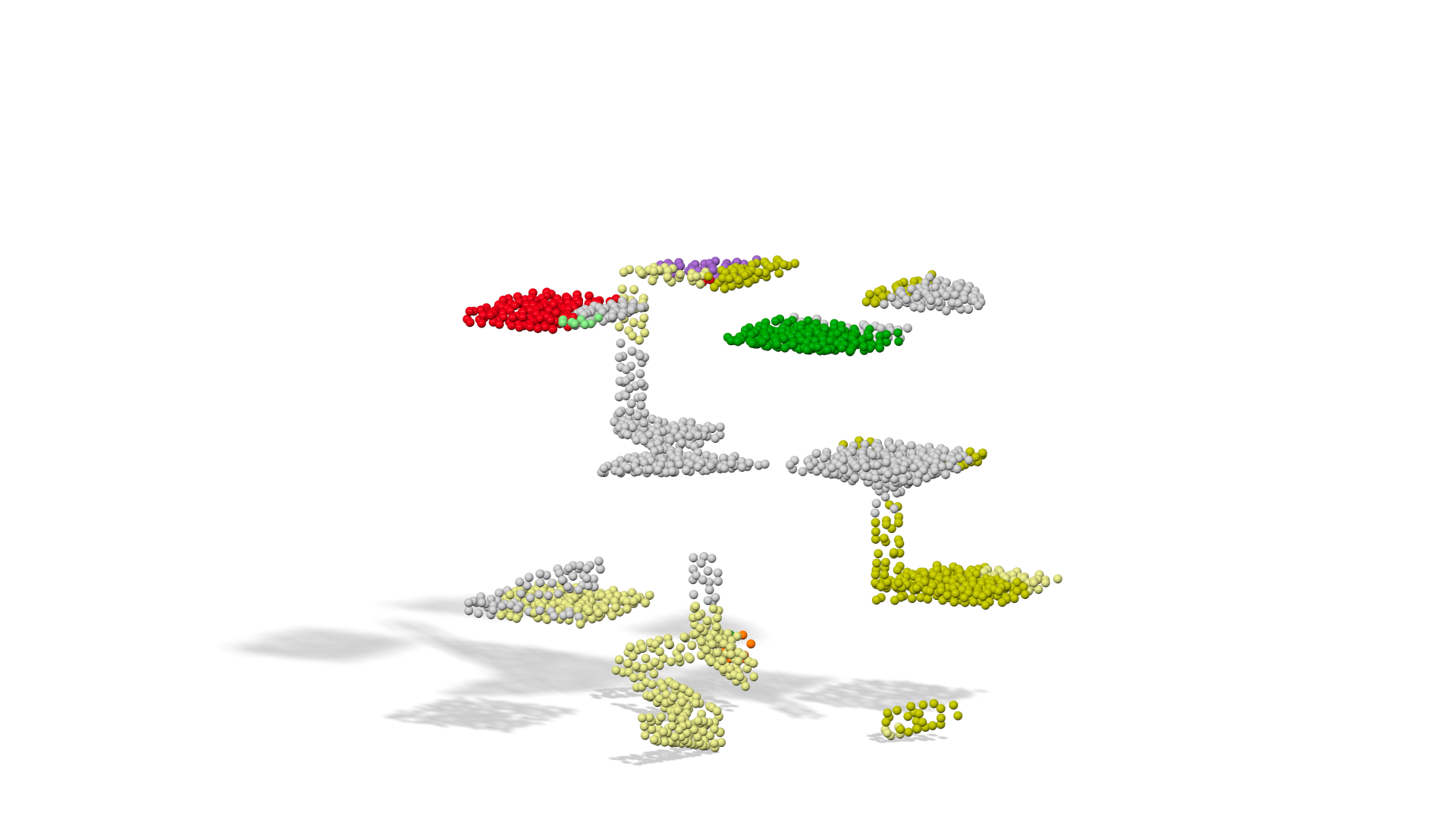}
\end{overpic}
\end{minipage}
\begin{minipage}{0.195\linewidth}
\centering
\begin{overpic}[width=\linewidth, clip, trim=250 0 400 0]
    {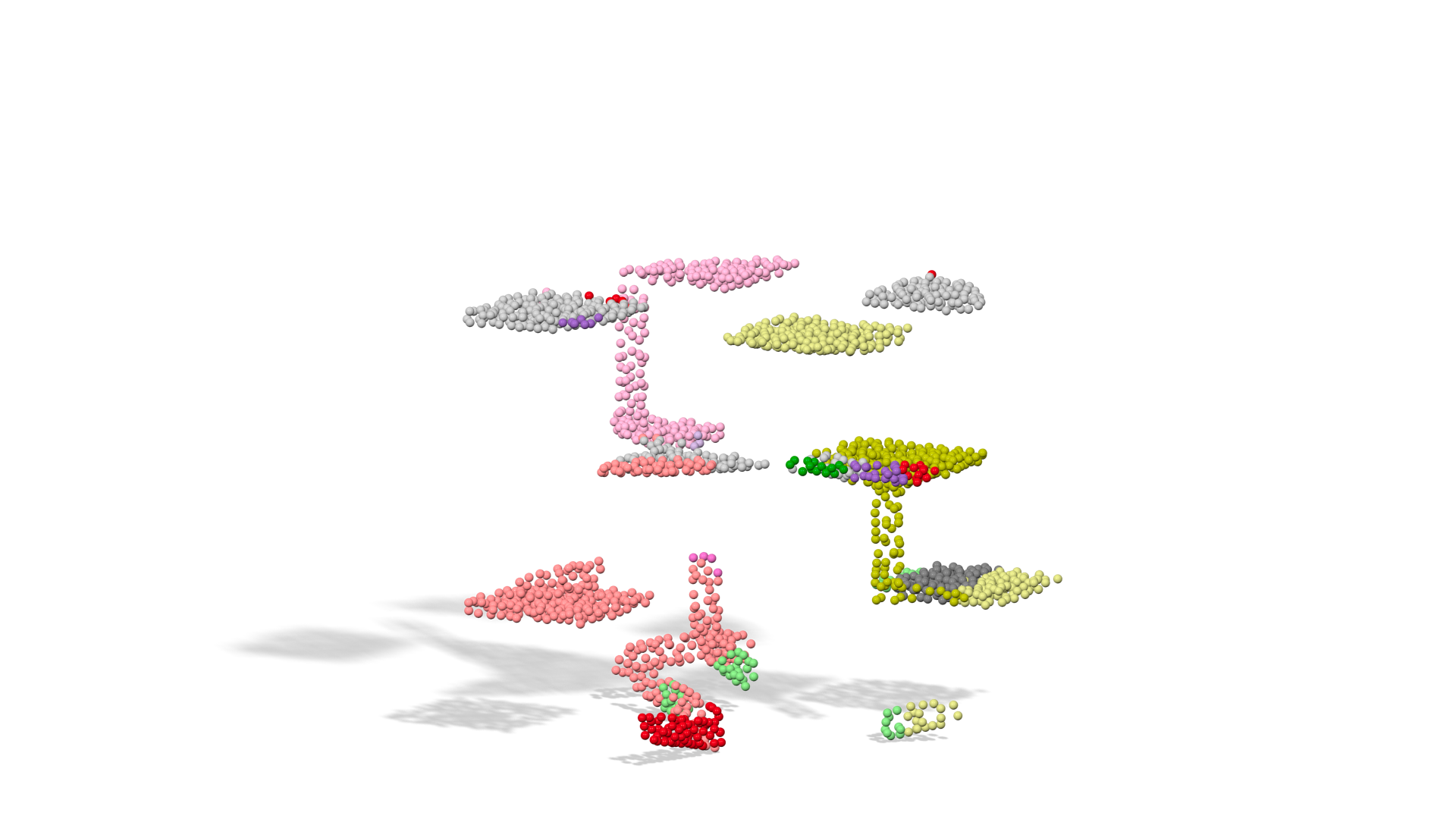}
\end{overpic}
\end{minipage}
\begin{minipage}{0.195\linewidth}
\centering
\begin{overpic}[width=\linewidth, clip, trim=250 0 400 0]
    {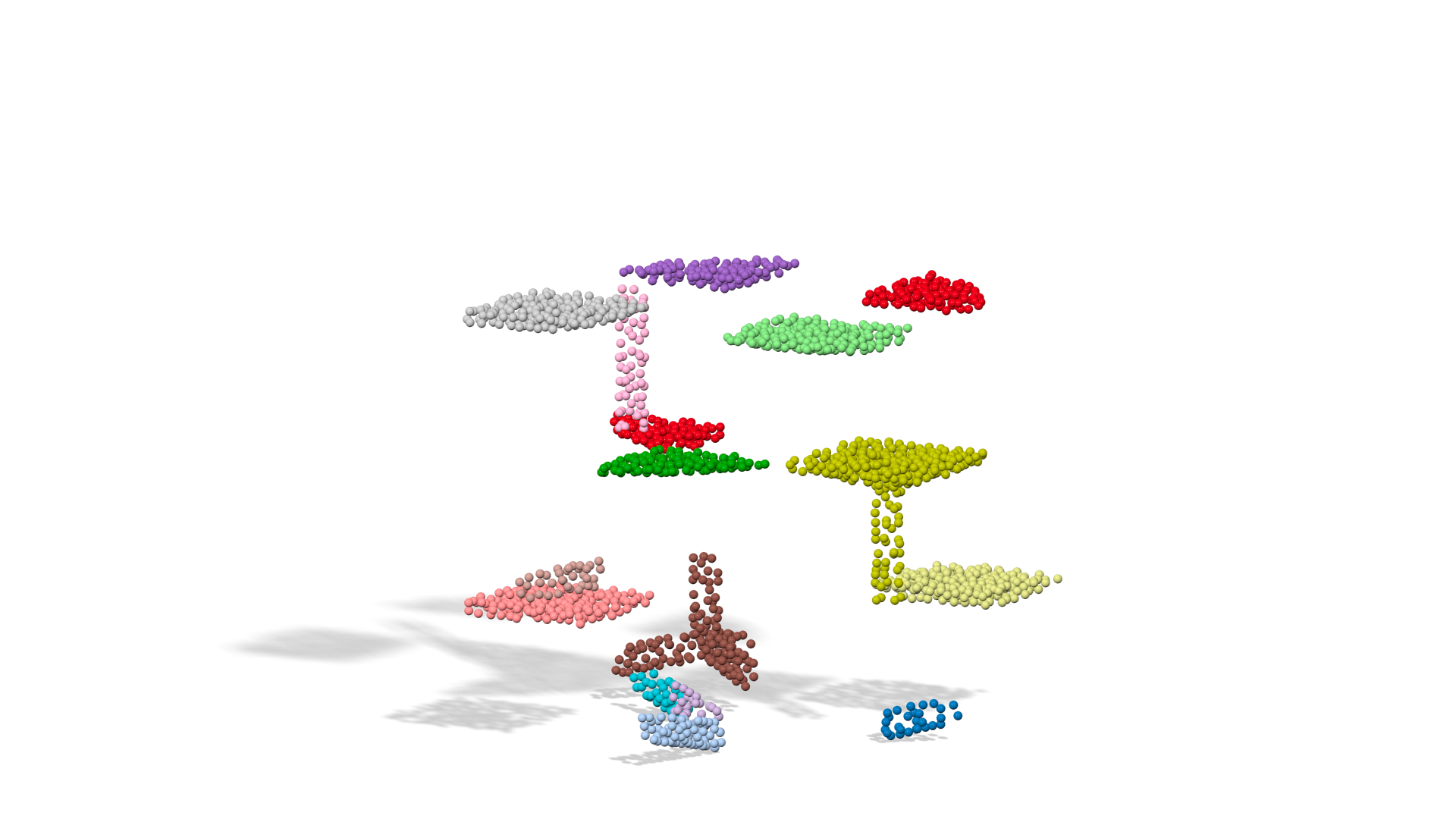}
\end{overpic}
\end{minipage} \\
\begin{minipage}{0.195\linewidth}
\centering
\begin{overpic}[width=\linewidth, clip, trim=400 0 350 0]
    {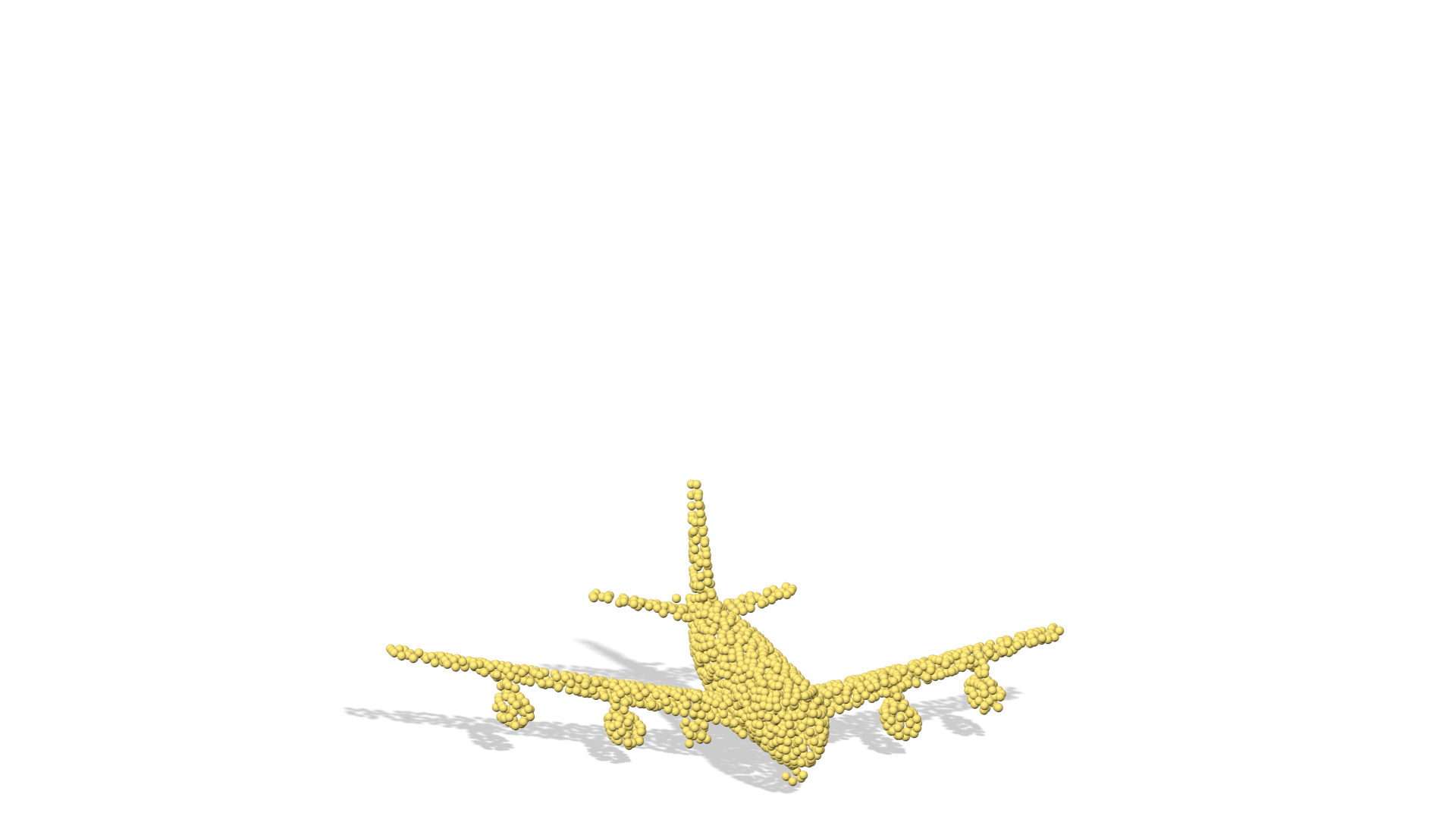}
\end{overpic}
Original
\end{minipage}
\begin{minipage}{0.195\linewidth}
\centering
\begin{overpic}[width=\linewidth, clip, trim=400 0 350 0]
    {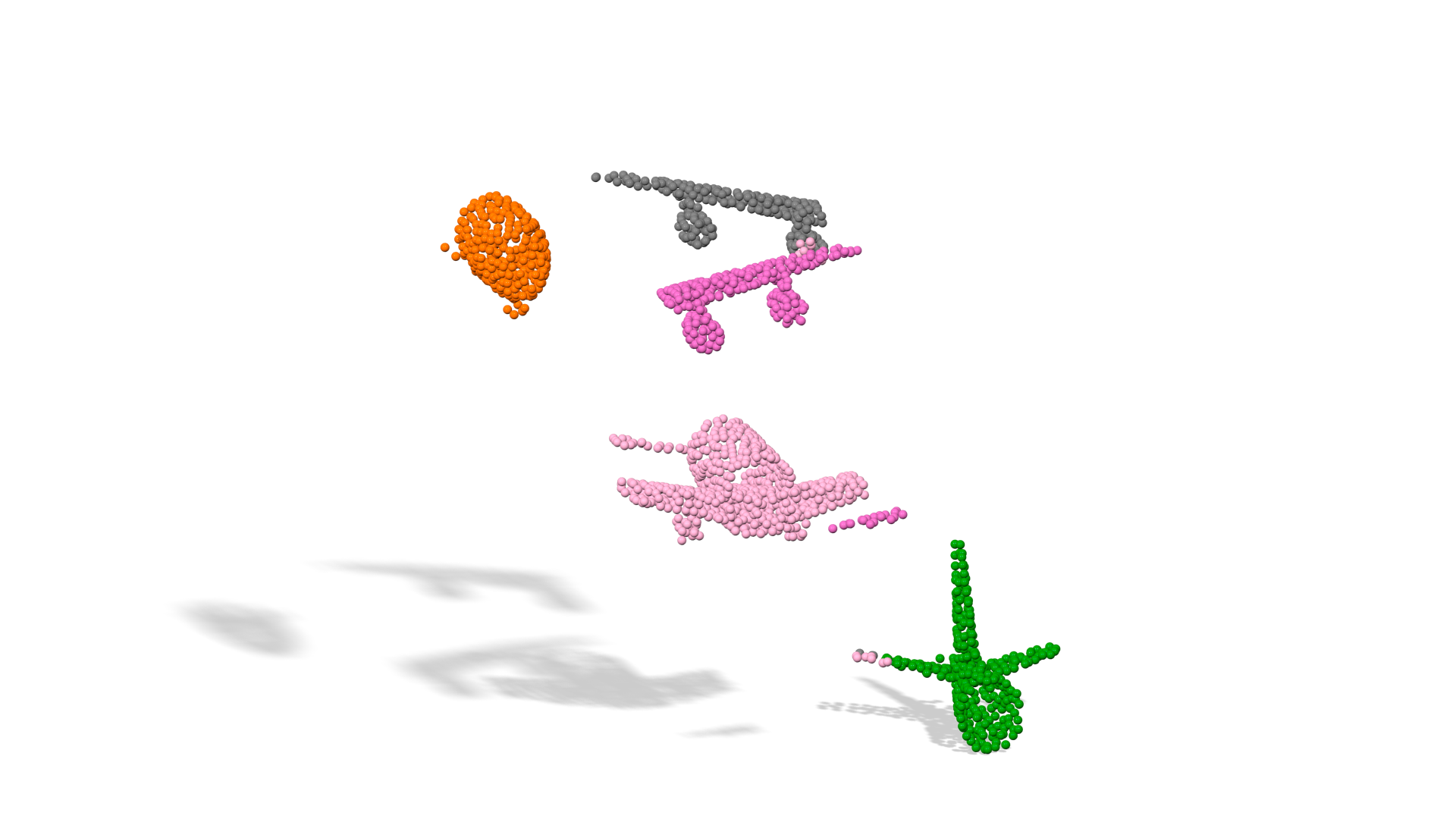}
    \put(-5, -19){\rule{.5pt}{190pt}}
\end{overpic}
Epoch = 0
\end{minipage}
\begin{minipage}{0.195\linewidth}
\centering
\begin{overpic}[width=\linewidth, clip, trim=400 0 350 0]
    {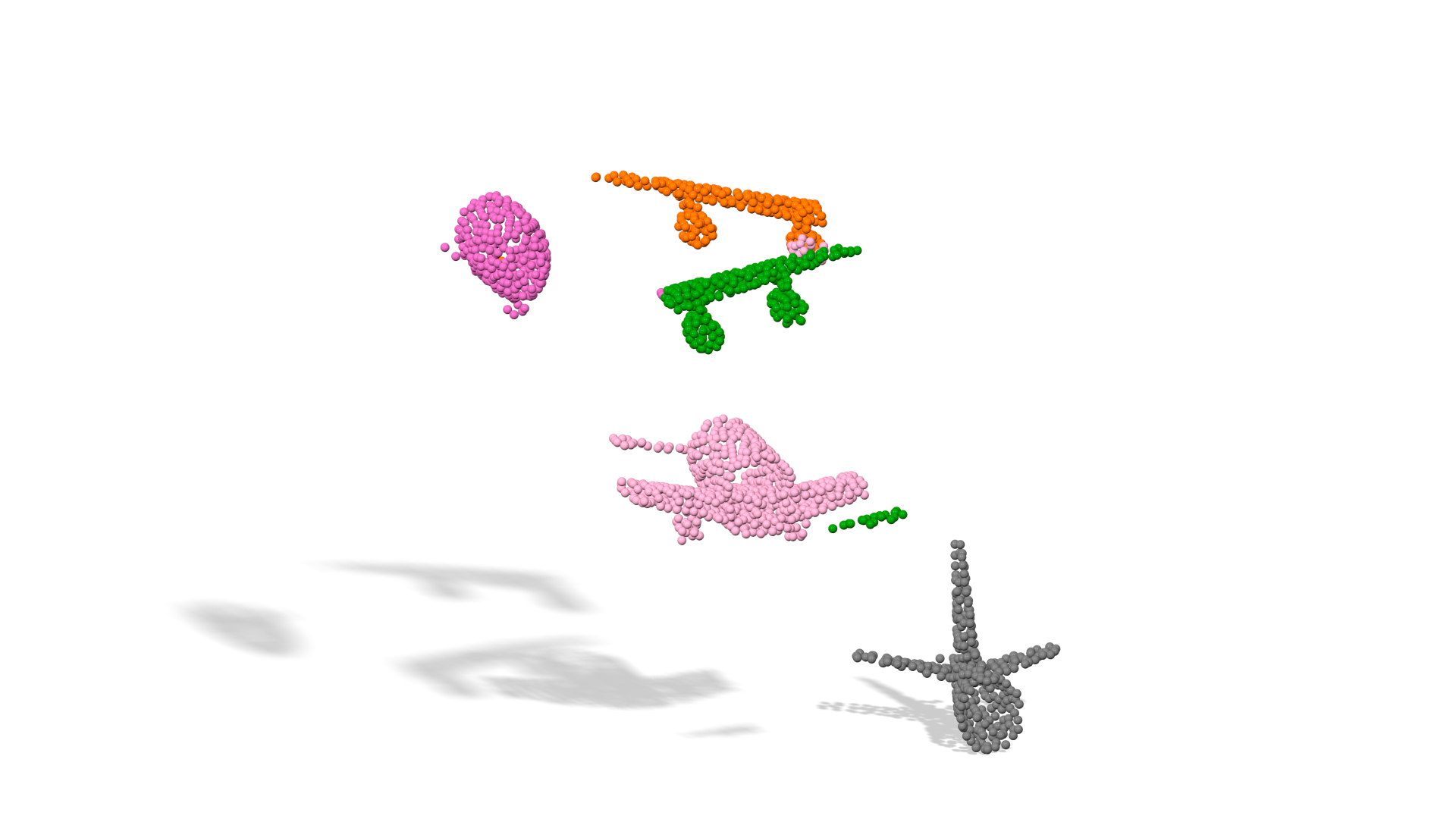}
\end{overpic}
Epoch = 10
\end{minipage}
\begin{minipage}{0.195\linewidth}
\centering
\begin{overpic}[width=\linewidth, clip, trim=400 0 350 0]
    {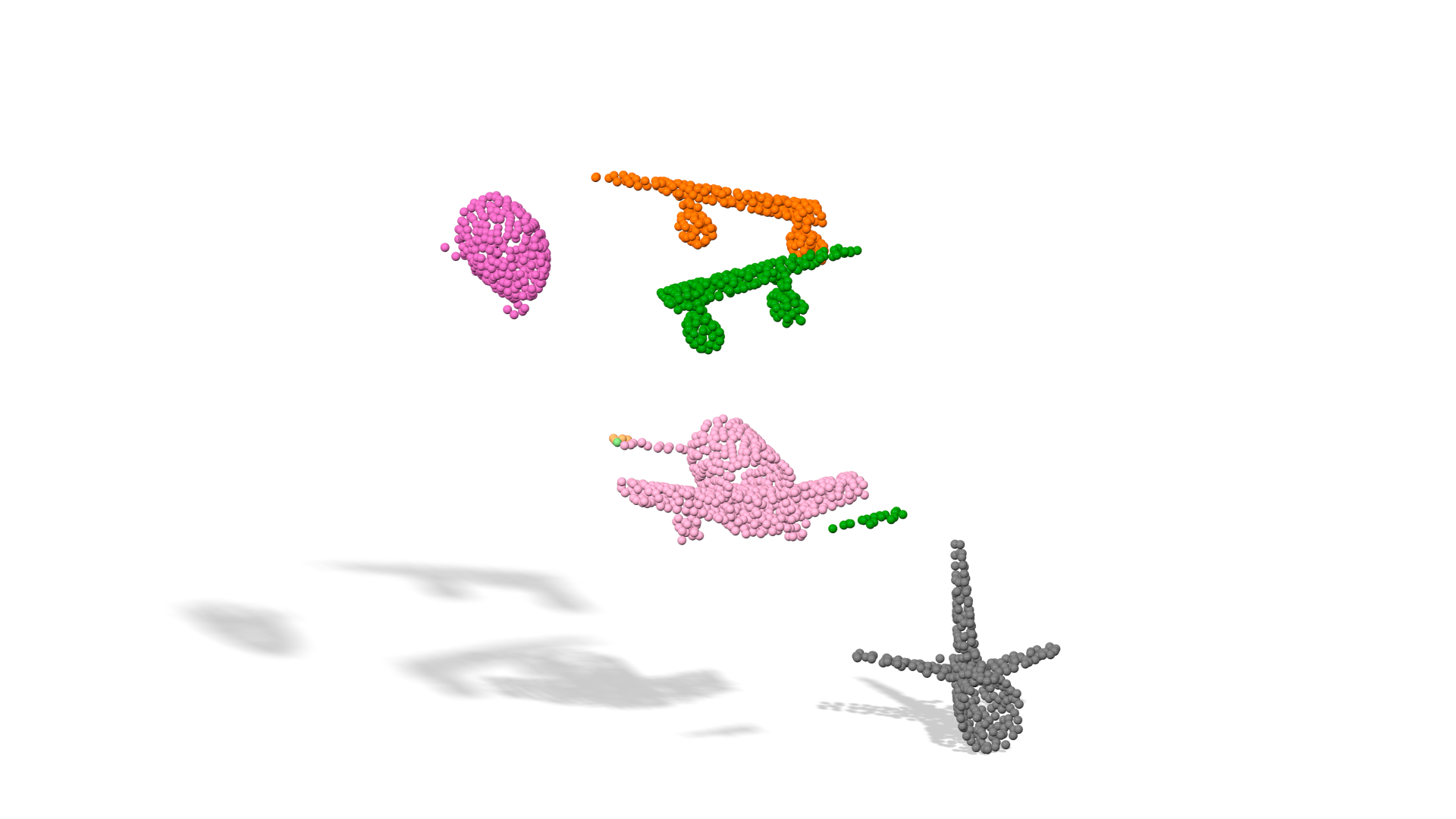}
\end{overpic}
Epoch = 20
\end{minipage}
\begin{minipage}{0.195\linewidth}
\vspace{-0.5mm}
\begin{overpic}[width=\linewidth, clip, trim=400 0 350 0]
    {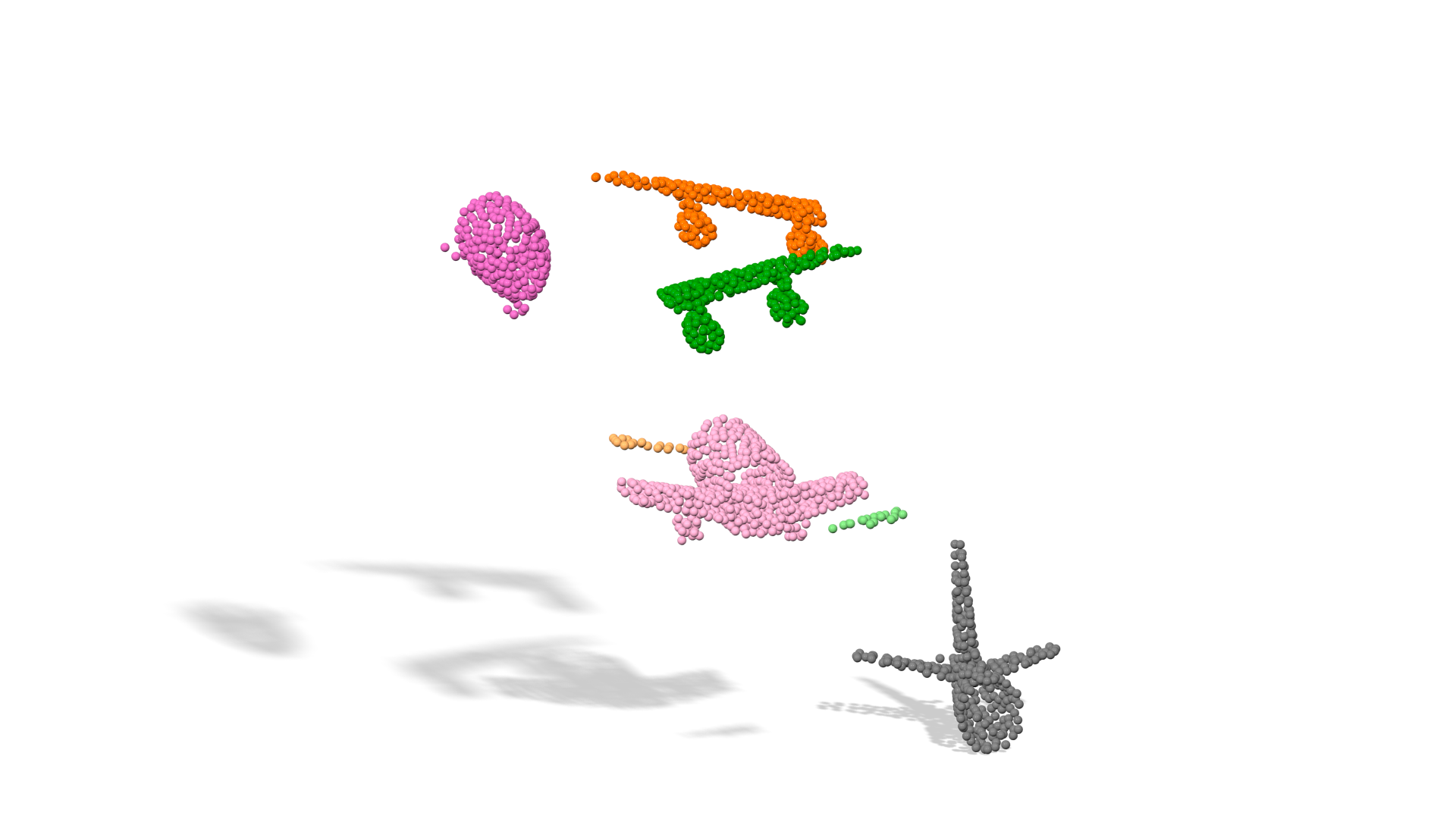}
    \put(-5, -19){\rule{.5pt}{190pt}}
\end{overpic}
Ground-truth
\end{minipage} \\
\smallskip
\caption{Visualization of 3D puzzle progressively solved by our model. 
\emph{First row}: the shuffled parts of the lamp are almost uniformly wrong at the beginning. The voxels are then identified during training. 
\emph{Second row}: annotating the correct voxel appears particularly difficult for this table due to multiple similar subparts. \emph{Third row}: for the airplane it seems easy to get coherent predictions but the voxel identity is initially mistaken and progressively corrected in the following epochs}
\label{fig:puzzle_solving}
\vspace{-3mm}
\end{figure*}

For a qualitative analysis of the learning process we show in 
Figure \ref{fig:puzzle_solving} how the puzzle is progressively solved at different epochs. Initially the annotation is quite confused, but then the model quickly improves by identifying to which decomposed voxel the sample part should belong to, and colors the points accordingly.

\section{Conclusions}
\label{sec:concl}
In this work we investigated how to deal with 3D labeled and unlabeled data possibly coming from different domains and with data annotation scarcity. To tackle these real world challenges  
we proposed a multi-task approach that combines supervised and self-supervised learning and showed with an extensive evaluation that it produces the new state-of-the-art for both shape classification and part segmentation on cross-domain and few-shot real world settings.
We see this work as a first exciting step towards a new family of methods better able to generalize and adapt to novel testing conditions for 3D point clouds. Our choice of the specific self-supervised task of solving 3D puzzle is indeed just one of the many possible that deserve attention for future work.

\clearpage
\bibliographystyle{splncs04}
\bibliography{biblio} 

\begin{thebibliography}{10}
\providecommand{\url}[1]{\texttt{#1}}
\providecommand{\urlprefix}{URL }
\providecommand{\doi}[1]{https://doi.org/#1}

\bibitem{achituve2020selfsupervised}
Achituve, I., Maron, H., Chechik, G.: Self-supervised learning for domain
  adaptation on point-clouds. Preprint ArXiv:2003.12641  (2020)

\bibitem{hdivergence}
Ben-David, S., Blitzer, J., Crammer, K., Kulesza, A., Pereira, F., Vaughan, J.:
  A theory of learning from different domains. Machine Learning  \textbf{79},
  151--175 (2010)

\bibitem{ben20183dmfv}
Ben-Shabat, Y., Lindenbaum, M., Fischer, A.: 3dmfv: Three-dimensional point
  cloud classification in real-time using convolutional neural networks. IEEE
  Robotics and Automation Letters  \textbf{3}(4),  3145--3152 (2018)

\bibitem{lscnn_2015}
Boscaini, D., Masci, J., Melzi, S., Bronstein, M.M., Castellani, U.,
  Vandergheynst, P.: Learning class-specific descriptors for deformable shapes
  using localized spectral convolutional networks. Comput. Graph. Forum
  \textbf{34},  13--23 (2015)

\bibitem{Bousmalis:DSN:NIPS16}
Bousmalis, K., Trigeorgis, G., Silberman, N., Krishnan, D., Erhan, D.: {Domain
  Separation Networks}. In: NIPS (2016)

\bibitem{Carlucci_2019_CVPR}
Carlucci, F.M., D'Innocente, A., Bucci, S., Caputo, B., Tommasi, T.: Domain
  generalization by solving jigsaw puzzles. In: CVPR (2019)

\bibitem{carlucci2017auto}
Carlucci, F.M., Porzi, L., Caputo, B., Ricci, E., Rota~Bul{\`o}, S.: Autodial:
  Automatic domain alignment layers. In: ICCV (2017)

\bibitem{ADAGE19}
Carlucci, F.M., Russo, P., Tommasi, T., Caputo, B.: Hallucinating agnostic
  images to generalize across domains. In: TASK-CV Workshop at ICCV (2019)

\bibitem{shapenet_dataset}
Chang, A.X., Funkhouser, T., Guibas, L., Hanrahan, P., Huang, Q., Li, Z.,
  Savarese, S., Savva, M., Song, S., Su, H., Xiao, J., Yi, L., Yu, F.:
  {ShapeNet: An Information-Rich 3D Model Repository}. In: Preprint
  ArXiv:1512.03012 (2015)

\bibitem{dai2017scannet}
Dai, A., Chang, A.X., Savva, M., Halber, M., Funkhouser, T., Nie{\ss}ner, M.:
  Scannet: Richly-annotated 3d reconstructions of indoor scenes. In: CVPR
  (2017)

\bibitem{doersch2015unsupervised}
Doersch, C., Gupta, A., Efros, A.A.: Unsupervised visual representation
  learning by context prediction. In: ICCV (2015)

\bibitem{Ganin:DANN:JMLR16}
Ganin, Y., Ustinova, E., Ajakan, H., Germain, P., Larochelle, H., Laviolette,
  F., Marchand, M., Lempitsky, V.: Domain-adversarial training of neural
  networks. J. Mach. Learn. Res.  \textbf{17}(1),  2096--2030 (2016)

\bibitem{DGautoencoders}
Ghifary, M., Kleijn, W.B., Zhang, M., Balduzzi, D.: Domain generalization for
  object recognition with multi-task autoencoders. In: ICCV (2015)

\bibitem{gidaris2018unsupervised}
Gidaris, S., Singh, P., Komodakis, N.: Unsupervised representation learning by
  predicting image rotations. In: ICLR (2018)

\bibitem{Han_2019_ICCV}
Han, Z., Wang, X., Liu, Y.S., Zwicker, M.: Multi-angle point cloud-vae:
  Unsupervised feature learning for 3d point clouds from multiple angles by
  joint self-reconstruction and half-to-half prediction. In: ICCV (2019)

\bibitem{Hassani_2019_ICCV}
Hassani, K., Haley, M.: Unsupervised multi-task feature learning on point
  clouds. In: ICCV (2019)

\bibitem{cycada}
Hoffman, J., Tzeng, E., Park, T., Zhu, J.Y., Isola, P., Saenko, K., Efros, A.,
  Darrell, T.: {C}y{CADA}: Cycle-consistent adversarial domain adaptation. In:
  ICML (2018)

\bibitem{Larsson_2017_CVPR}
Larsson, G., Maire, M., Shakhnarovich, G.: Colorization as a proxy task for
  visual understanding. In: CVPR (2017)

\bibitem{MLDG_AAA18}
Li, D., Yang, Y., Song, Y., Hospedales, T.M.: Learning to generalize:
  Meta-learning for domain generalization. In: AAAI (2018)

\bibitem{Li_2019_ICCV}
Li, D., Zhang, J., Yang, Y., Liu, C., Song, Y.Z., Hospedales, T.M.: Episodic
  training for domain generalization. In: ICCV (2019)

\bibitem{Li_2018_CVPR}
Li, H., Jialin~Pan, S., Wang, S., Kot, A.C.: Domain generalization with
  adversarial feature learning. In: CVPR (2018)

\bibitem{sonet_2018_CVPR}
Li, J., Chen, B.M., Hee~Lee, G.: So-net: Self-organizing network for point
  cloud analysis. In: CVPR (2018)

\bibitem{Li_2018_ECCV}
Li, Y., Tian, X., Gong, M., Liu, Y., Liu, T., Zhang, K., Tao, D.: Deep domain
  generalization via conditional invariant adversarial networks. In: ECCV
  (2018)

\bibitem{NIPS2018_7362}
Li, Y., Bu, R., Sun, M., Wu, W., Di, X., Chen, B.: Pointcnn: Convolution on
  x-transformed points. In: NIPS (2018)

\bibitem{Long:2015}
Long, M., Cao, Y., Wang, J., Jordan, M.I.: Learning transferable features with
  deep adaptation networks. In: ICML (2015)

\bibitem{LongZ0J17}
Long, M., Zhu, H., Wang, J., Jordan, M.I.: Deep transfer learning with joint
  adaptation networks. In: ICML (2017)

\bibitem{mancini2018boosting}
Mancini, M., Porzi, L., Rota~Bul\`o, S., Caputo, B., Ricci, E.: Boosting domain
  adaptation by discovering latent domains. In: CVPR (2018)

\bibitem{monet_2017}
Monti, F., Boscaini, D., Masci, J., Rodol{\`{a}}, E., Svoboda, J., Bronstein,
  M.M.: Geometric deep learning on graphs and manifolds using mixture model
  cnns. In: CVPR (2017)

\bibitem{noroozi2016}
Noroozi, M., Favaro, P.: Unsupervised learning of visual representations by
  solving jigsaw puzzles. In: ECCV (2016)

\bibitem{pathakCVPR16context}
Pathak, D., Kr\"ahenb\"uhl, P., Donahue, J., Darrell, T., Efros, A.: Context
  encoders: Feature learning by inpainting. In: CVPR (2016)

\bibitem{pointnet_CVPR17}
Qi, C.R., Su, H., Mo, K., Guibas, L.J.: {PointNet: Deep Learning on Point Sets
  for 3D Classification and Segmentation}. In: CVPR (2017)

\bibitem{pointnet++_NIPS2017}
Qi, C.R., Yi, L., Su, H., Guibas, L.J.: {PointNet++: Deep Hierarchical Feature
  Learning on Point Sets in a Metric Space}. In: NIPS (2017)

\bibitem{pointdan10}
Qin, C., You, H., Wang, L., Kuo, C.C.J., Fu, Y.: Pointdan: A multi-scale 3d
  domain adaption network for point cloud representation. In: NIPS (2019)

\bibitem{russo17sbadagan}
Russo, P., Carlucci, F.M., Tommasi, T., Caputo, B.: From source to target and
  back: symmetric bi-directional adaptive gan. In: CVPR (2018)

\bibitem{saito2017maximum}
Saito, K., Watanabe, K., Ushiku, Y., Harada, T.: Maximum classifier discrepancy
  for unsupervised domain adaptation. CVPR  (2018)

\bibitem{saudersievers}
Sauder, J., Sievers, B.: Self-supervised deep learning on point clouds by
  reconstructing space. In: NIPS (2019)

\bibitem{DG_ICLR18}
Shankar, S., Piratla, V., Chakrabarti, S., Chaudhuri, S., Jyothi, P., Sarawagi,
  S.: Generalizing across domains via cross-gradient training. In: ICLR (2018)

\bibitem{dcoral}
Sun, B., Saenko, K.: Deep coral: Correlation alignment for deep domain
  adaptation. In: ECCV Workshops (2016)

\bibitem{Hoffman:Adda:CVPR17}
Tzeng, E., Hoffman, J., Darrell, T., Saenko, K.: Adversarial discriminative
  domain adaptation. In: CVPR (2017)

\bibitem{scanobjectnn_ICCV19}
Uy, M.A., Pham, Q.H., Hua, B.S., Nguyen, D.T., Yeung, S.K.: {Revisiting Point
  Cloud Classification: A New Benchmark Dataset and Classification Model on
  Real-World Data}. In: ICCV (2019)

\bibitem{Volpi_2018_NIPS}
Volpi, R., Namkoong, H., Sener, O., Duchi, J., Murino, V., Savarese, S.:
  Generalizing to unseen domains via adversarial data augmentation. In: NIPS
  (2018)

\bibitem{localspectral}
Wang, C., Samari, B., Siddiqi, K.: Local spectral graph convolution for point
  set feature learning. In: ECCV (2018)

\bibitem{dgcnn}
Wang, Y., Sun, Y., Liu, Z., Sarma, S.E., Bronstein, M.M., Solomon, J.M.:
  {Dynamic Graph CNN for Learning on Point Clouds}. ACM Trans. Graph.  (2019)

\bibitem{mn40_dataset}
Wu, Z., Song, S., Khosla, A., Yu, F., Zhang, L., Tang, X., Xiao, J.: {3D
  ShapeNets: A Deep Representation for Volumetric Shapes}. In: CVPR (2015)

\bibitem{lopez}
Xu, J., Xiao, L., L{\'{o}}pez, A.M.: Self-supervised domain adaptation for
  computer vision tasks. Preprint Arxiv:1907.10915  (2019)

\bibitem{Xu_2019_ICCV}
Xu, R., Li, G., Yang, J., Lin, L.: Larger norm more transferable: An adaptive
  feature norm approach for unsupervised domain adaptation. In: ICCV (2019)

\bibitem{xu2018spidercnn}
Xu, Y., Fan, T., Xu, M., Zeng, L., Qiao, Y.: Spidercnn: Deep learning on point
  sets with parameterized convolutional filters. In: ECCV (2018)

\bibitem{Yang_2018_CVPR}
Yang, Y., Feng, C., Shen, Y., Tian, D.: Foldingnet: Point cloud auto-encoder
  via deep grid deformation. In: CVPR (2018)

\bibitem{shapenetpart_dataset}
Yi, L., Kim, V.G., Ceylan, D., Shen, I.C., Yan, M., Su, H., Lu, C., Huang, Q.,
  Sheffer, A., Guibas, L.: {A Scalable Active Framework for Region Annotation
  in 3D Shape Collections}. In: SIGGRAPH Asia (2016)

\bibitem{Yi_2017_CVPR}
Yi, L., Su, H., Guo, X., Guibas, L.J.: Syncspeccnn: Synchronized spectral cnn
  for 3d shape segmentation. In: CVPR (2017)

\bibitem{Zhai_2019_ICCV}
Zhai, X., Oliver, A., Kolesnikov, A., Beyer, L.: S4l: Self-supervised
  semi-supervised learning. In: ICCV (2019)

\bibitem{zhang2018mixup}
Zhang, H., Cisse, M., Dauphin, Y.N., Lopez-Paz, D.: mixup: Beyond empirical
  risk minimization. In: ICLR (2018)

\bibitem{3dworkshopcvpr}
Zhang, L., Zhu, Z.: Unsupervised feature learning for point cloud by
  contrasting and clustering with graph convolutional neural network. In: CVPR
  Workshop (2019)

\bibitem{zhang2016colorful}
Zhang, R., Isola, P., Efros, A.A.: Colorful image colorization. In: ECCV (2016)

\bibitem{Zhao_2019_CVPR}
Zhao, Y., Birdal, T., Deng, H., Tombari, F.: 3d point capsule networks. In:
  CVPR (2019)

\end{thebibliography}

\section*{Appendix}
\appendix
\section{Relation to other Puzzle Solvers}
We designed our puzzle solver with the aim of making it easily compatible with the main supervised objective in a multi-task model specifically tailored for 3D problems.
Indeed our learning architecture is trained by jointly optimizing both the puzzle and the main supervised task. This makes our approach different from the recently published work \cite{saudersievers} that discusses a 3D puzzle task whose self-supervised model is learned in isolation and only in a second phase transferred to a down-stream task. On the other hand, a related work exploiting supervised and self-supervised multi-task learning is \cite{Carlucci_2019_CVPR} where, however, the puzzle task is defined in 2D with a completely different logic with respect to that proposed in our work. Specifically in \cite{Carlucci_2019_CVPR} the whole puzzled sample is described by a single index which identifies the voxels' permutation and the puzzle task is formalized as a classification problem to predict that index. This implementation does not have a straightforward extension to 3D because it does not deal properly with the non-Euclidean nature of point clouds, and  makes the puzzle particularly difficult in case of empty voxels. A practical case would be that of having $l=3$ or $l=4$ as discussed in sec. \ref{subsection:ablation}, where among the large number of voxels, only a limited amount will contain some point.
In a preliminary set of experiments we tested this global classification strategy observing poor results. This happens because the self-supervised task is not able to provide local auxiliary information and the multi-task model does not show any advantage wrt the single-task supervised baseline.
Our model is instead tailored for 3D problems:  we assign a label to each object voxel, 
which is inherited by the vertices of the point cloud contained in that voxel.
After shuffling, the puzzle solver performs a per-point voxel label prediction. Since the focus is on the points and not on the voxels, the issue of empty voxels does not affect our approach.

\section{Further Experimental Comparisons in the DA Setting}
In sec. \ref{subsection:classification} we discussed the experimental comparison of our approach against PointDAN~\cite{pointdan10}, which is, at the time of writing, the only publication focusing on 3D pointcloud classification across domains. We extend here the evaluation on the dataset proposed in \cite{pointdan10}, built upon 10 overlapping object classes between ModelNet-10 (M), ShapeNet (S) and ScanNet~\cite{dai2017scannet} (S*). The first two are synthetic data collections, while the last is a real-world dataset from RGB-D scans.

We consider each 3D object aligned with respect to the upward direction and unit-cube scaled. During training we perform the standard data augmentation described in previous experiments: random angle y-axis rotation and random vertex jittering drawn from $\mathcal{N}(0, 0.01)$. We also compare our result with a very recent pre-print \cite{achituve2020selfsupervised} which presents a domain adaptation approach based on the combination of region reconstruction (\emph{RegRec}) and a training procedure based on the Mixup method \cite{zhang2018mixup} (Point Cloud Mixup, \emph{PCM}). The first is indeed a self-supervised strategy, while the second is a training procedure that exploits linear combination of samples.

As shown in Table~\ref{tab:evaluation_pointdan}, even in this setting our multi-task approach outperforms PointDAN. With respect to the method in \cite{achituve2020selfsupervised}, the direct comparison with RegRec shows that this self-supervised reconstruction task is less powerful with respect to our 3D puzzle solver for adaptation. Moreover, the results seem quite unstable as shown by the large standard deviation for the $M\to S^*$ and $S^*\to M$ cases. The only advantage is provided by the PCM method for some of the domain pairs. Still, on average the results remain in favour of our method.

\begin{table*}[tb!]
    \centering
    \caption{Cross-Domain accuracy (\%) on PointDAN~\cite{pointdan10} dataset. We compare Our approach against PointDAN and the method proposed in the recent pre-print \cite{achituve2020selfsupervised} on top of the same PointNet~\cite{pointnet_CVPR17} feature extractor. We refer to ModelNet-10 as \emph{M}, ShapeNet-10 as \emph{S} and ScanNet-10 as \emph{S*}}
    \resizebox{\textwidth}{!}{
    \begin{tabular}{l@{~~}|@{~}c@{~~}c@{~~}c@{~~}c@{~~}c@{~~}c@{~~}|@{~}c}
        \hline
             Method & M$\to$S & M$\to$S* & S$\to$M & S$\to$S* & S*$\to$M & S*$\to$S & Avg.\\
        \hline
            PointDAN~\cite{pointdan10} & 80.2$\pm$0.8 & 45.3$\pm$2.0 & 71.2$\pm$3.0 & 46.9$\pm$3.3 & 59.8$\pm$2.3 & 66.2$\pm$4.8 & 61.6\\
            RegRec~\cite{achituve2020selfsupervised} & 80.0$\pm$0.6 & 46.0$\pm$5.7 & 68.5$\pm$4.8 & 41.7$\pm$1.9 & 63.0$\pm$6.7 & 68.2$\pm$1.1 & 61.2\\
            RegRec + PCM~\cite{achituve2020selfsupervised} & 81.1$\pm$1.1 & \textbf{50.3$\pm$2.0} & 54.3$\pm$0.3 & \textbf{52.8$\pm$2.0} & 54.0$\pm$5.5 & \textbf{69.0$\pm$0.9} & 60.3\\
            Our & \textbf{81.6$\pm$0.6} & 49.7$\pm$1.4 & \textbf{73.6$\pm$0.5} & 41.9$\pm$0.9 & \textbf{65.9$\pm$0.7} & 68.1$\pm$1.6 & \textbf{63.5}\\
        \hline
    \end{tabular}
    }
    \label{tab:evaluation_pointdan}
\end{table*}

\end{document}